\documentclass[lettersize,journal]{IEEEtran}
\usepackage{amsmath,amsfonts}
\usepackage{algorithmic}
\usepackage{algorithm}
\usepackage{array}
\usepackage[caption=false,font=normalsize,labelfont=sf,textfont=sf]{subfig}
\usepackage{textcomp}
\usepackage{stfloats}
\usepackage{url}
\usepackage{verbatim}
\usepackage{graphicx}
\usepackage{cite}
\usepackage{amsthm,amssymb}
\usepackage{mathrsfs}
\usepackage[table]{xcolor}
\usepackage{bbm}
\begin{document}

\title{Renormalized Connection for Scale-preferred Object Detection in Satellite Imagery}

\author{Fan Zhang, Lingling Li, \IEEEmembership{Senior Member, IEEE}, Licheng Jiao, \IEEEmembership{Fellow, IEEE}, Xu Liu, \IEEEmembership{Senior Member, IEEE}, Fang Liu, \IEEEmembership{Senior Member, IEEE}, Shuyuan Yang, \IEEEmembership{Senior Member, IEEE}, Biao Hou, \IEEEmembership{Senior Member, IEEE}
\thanks{The authors are with the Key Laboratory of Intelligent Perception and Image Understanding of Ministry of Education, International Research Center for Intelligent Perception and Computation, School of Artificial Intelligence, Xidian University, Xi’an 710071, China (e-mail: lchjiao@mail.xidian.edu.cn).}
}

\markboth{Published in IEEE TRANSACTIONS ON GEOSCIENCE AND REMOTE SENSING (08/2024)}%
{Shell \MakeLowercase{\textit{et al.}}: A Sample Article Using IEEEtran.cls for IEEE Journals}


\maketitle
\begin{abstract}
Satellite imagery, due to its long-range imaging, brings with it a variety of scale-preferred tasks, such as the detection of tiny/small objects, making the precise localization and detection of small objects of interest a challenging task. In this article, we design a Knowledge Discovery Network (KDN) to implement the renormalization group theory in terms of efficient feature extraction. Renormalized connection (RC) on the KDN enables ``synergistic focusing'' of multi-scale features. Based on our observations of KDN, we abstract a class of renormalized connections with different connection strengths, called $n21$C, and generalize it to FPN-based multi-branch detectors. In a series of FPN experiments on the scale-preferred tasks, we found that the ``divide-and-conquer'' idea of FPN severely hampers the detector's learning in the right direction due to the large number of large-scale negative samples and interference from background noise. Moreover, these negative samples cannot be eliminated by the focal loss function. The Renormalized Connections extends the multi-level feature's ``divide-and-conquer'' mechanism of the FPN-based detectors to a wide range of scale-preferred tasks, and enables synergistic effects of multi-level features on the specific learning goal. In addition, interference activations in two aspects are greatly reduced and the detector learns in a more correct direction. Extensive experiments of 17 well-designed detection architectures embedded with $n21$Cs on five different levels of scale-preferred tasks validate the effectiveness and efficiency of the Renormalized Connections. Especially the simplest linear form of RC --- E421C performs well in all tasks and it satisfies the scaling property of renormalization group theory. All experiments can be trained and tested on a graphics card with 8GB of video memory, which greatly enhances the applicability of our methodology. We hope that our approach will transfer a large number of well-designed detectors from the computer vision community to the remote sensing community. Datasets and codes will be available at: https://github.com/rabbitme/.
\end{abstract}


\section{Introduction}
\label{intro}

\begin{figure}[!t]
\centering
\includegraphics[width=3.5in]{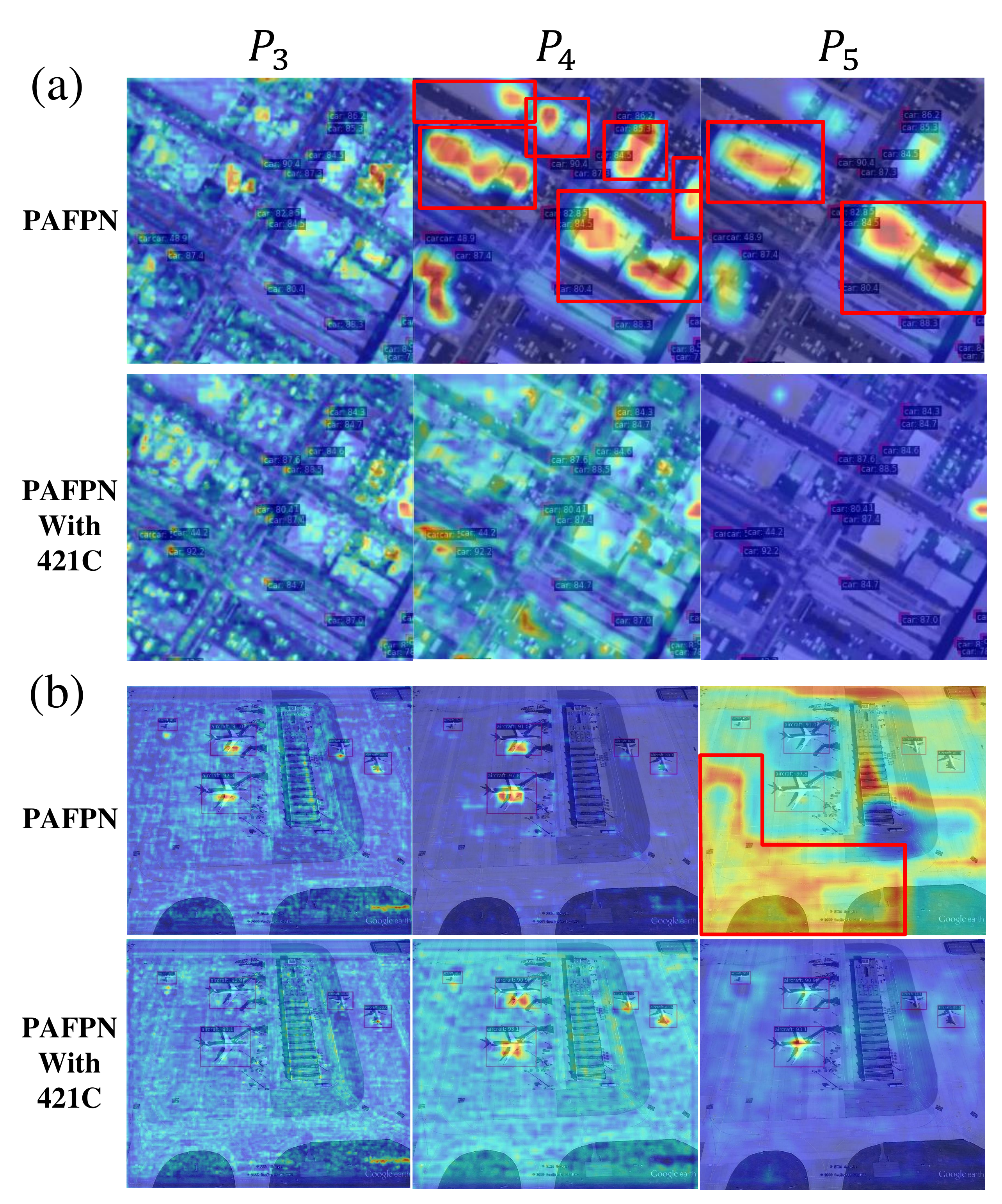}
\caption{The saliency maps of YOLOv8-PAFPN describe the importance of features. The FPN series of extractors with focal loss are unable to adapt to difficult scale-preferred tasks. The initial ``divide-and-conquer'' mechanism for different pyramid levels brings a large number of interfering activations (areas with red activation values) for tiny object detection tasks. Fig. \ref{heatmap_comparison} shows the saliency maps of the three pyramid level feature activations of the feature extractor PAFPN for two types of scale preference tasks. (a) shows a purely tiny object detection task, we can see that the $P_4$ and $P_5$ level feature maps highlight the blockbuster background objects that have similar visual appearances (color and shape) but with different scales and object classes (in red boxes). This will greatly overwhelm the detection focus of $P_3$. (b) shows a small object detection task, where the $P_3$ and $P_4$ level all focus on the same-sized objects which ought to be detected by $P_4$. The $P_5$ level, however, is more concerned with the large irrelevant regions in the background. All the observations suggest that the signal-to-noise ratio of the FPNs used for the scale-preferred tasks is lower than the scale-diversified tasks because the huge amount of interfering activations overwhelmingly prevent the previous levels ((a) $P_3$, (b) $P_3$\&$P_4$) from focusing on small objects as the objective.}
\label{heatmap_comparison}
\end{figure}

\IEEEPARstart{W}{ith} the rapid advancement of deep learning technology and hardware, the application scope of intelligent interpretation of satellite imagery has gradually expanded. In the field of remote sensing, the accuracy of automatic detection of large objects has become comparable to that of human experts \cite{10402099}\cite{jiao2019survey}, and has been applied to aspects of people's production and life. However, there are a large number of small objects in satellite imagery \cite{10168277}\cite{9208789}, and there is still room to improve the accuracy of small object detection. In generic object detection community, there are a huge number of well-designed algorithms including higher capacity backbone networks \cite{he2016deep}\cite{huang2017densely}\cite{9879380}, feature extractors with rich feature space \cite{fpn}\cite{liu2018path}\cite{ghiasi2019fpn}\cite{tan2020efficientdet}\cite{f}, aligned head network \cite{9710724} to cope with performance enhancement problems. Compared with the scale-diversified object detection tasks, small object detection tasks can be seen as a scale preference problem.

Recently, the remote sensing community has proposed well-designed tasks and solutions to concentrate on small object detection \cite{9143165}. For accurate object recognition, mining useful features in a huge space is an important direction. Jia et al. \cite{jia2013feature} investigated in detail a series of feature mining methods for hyperspectral image classification, including feature selection and feature extraction methods. Chen et al. \cite{chen2016deep} presented a regularized deep feature extraction (FE) method and verified its validity. Ma et al. \cite{ma2023multiscale} designed a spectral-spatial feature mining framework for noise-robust feature extraction. Context features are another crucial piece of information. Li et al. \cite{10509806} raised an SDFF architecture that performs inter-layer feature fusion to incorporate multi-scale context in the dynamic image pyramid of the large-size image. Luo et al. \cite{luo2024pointobb} first used a single Point-based OBB generation method for oriented object detection. Inspired by these effective works, we design feature mining methods to tackle difficult scale preference tasks. 

For scale related problems, the most popular feature extraction method is FPN \cite{fpn}, which can simultaneously detect different scale objects in an end-to-end manner. The core design idea of FPN is ``divide-and-conquer'', that is, each feature level in the pyramid performs its own function, to detect objects within a certain scale range. Based on this, PAFPN \cite{liu2018path} extended the number of connection layers of the feature extractor by downsampling the largest feature map output from the FPN. It is straightforward to use FPN-based detectors to tackle scale-preferred problems. However, when we study scale-preferred tasks, such as the tiny/small object detection tasks (Fig. \ref{heatmap_comparison}), the direct use of the FPN detector leads to serious problems with interfering activations, both in terms of large numbers of negative samples and background noises, caused by the ``divide-and-conquer'' idea and the multi-subtask parallel learning of different feature levels of FPN. 

Notably, Fig. \ref{heatmap_comparison} shows the saliency map of the one-stage detector YOLOv8-PAFPN coupled with focal loss function \cite{yolov8} after 500 epochs of training. As we know, the focal loss \cite{lin2017focal} aims to address the foreground-background class imbalance problem of one-stage detectors on scale-diversified datasets. However, on scale-preferred datasets, a large number of negative samples are actively generated which produce large activation values that cannot be eliminated during training. These two aspects of the observations suggest that there is still a noticeable problem in scale-preferred tasks. That is, the high-level features of FPN propose a large number of negative samples, and the focal loss cannot eliminate these interfering activations during training.

In small object detection tasks, if the interfering activations in the large object detection branch are eliminated, there will be no interfering gradient information when running backpropagation. The learning process will then be redirected in the right direction. Therefore, we aim to address the problem that branches assigned fewer positive samples produce interference activations (the distribution of activation values can be visualized from the saliency maps) in tasks with only small objects and, more generally, in tasks with scale preference. If this problem is solved, then a large number of methods designed for tasks with diverse scales can be used to solve the scale-preferred problem, thus greatly expanding the set of solutions to various scale-preferred problems.  

In this article, because the focal loss cannot eliminate the actively generated large number of negative samples by FPN's $P_5$ level on scale-preferred tasks, we consider using renormalization group theory to guide the detector to focus on hard samples. At first, we design a Knowledge Discovery Network (KDN) on a single-branch detector to find the relationship between different stage features in the feature extractor and the function of each stage for the ultimate detection task. Without the need to design a novel positive-negative sample assignment method for the scale-preferred task, we start from the network structure design (KDN) and bridge a renormalized connection between the feature extractor and the detection head, directing the multi-level outputs of the feature extractor to be correctly used for the feature supply of the detection head. 

After a series of experiments of KDN on different small object detection datasets, we find that although the stages in ResNet, except the $C_3$ stage, are not directly involved in the detection task, each stage produces different levels of features that are both independent of and complementary to each other. This is consistent with the idea that different levels of features have their own roles in the FPN. Inspired by this observation, we generalize the renormalized connections obtained by KDN to the more general case of FPN-based multi-branch detectors. To mathematically abstract this renormalized connection method, we construct a set of renormalized coefficients, called connection strengths, for a set of mutually independent bases on the feature space generated by the feature extractor. We name this type of Renormalized Connection according to the coefficient ratios, as $n21$C. Among them, the simplest uniformly renormalized connection coefficients conform to the scaling property in renormalization group theory, with coefficient ratios of 421, denoted by 421C. 

The $n21$Cs represent a class of renormalized connections in which the coefficient ratios can be varied according to the task requirements, and the optimal coefficient ratios for each specific task can be found by a grid search or a random search. When used for tasks with specific scale preferences, we find analytically and experimentally in this article that very small, small, and scale-diversified tasks can use the 421 coefficient ratios to renormalize the connections of three or more different detection branches. The 421Cs eliminate interference activations on the large object detection branch and achieve the goal of detection branches not interfering with each other and learning collaboratively on the task of focusing on detecting small objects. The RCs extend the idea of the multi-level feature's ``divide-and-conquer'' mechanism of the FPN-based detectors to the scale-preferred tasks and enable synergistic effects of multi-level features on the specific learning goals. It greatly eliminates interfering activations from the branches assigned fewer positive samples of the FPN that generate a large number of negative samples and background noise before they are fed into the detection network. The $n21$Cs assign appropriate subtasks to multiple detection branches for scale-preferred tasks. In a word, Renormalized Connections in multi-branch detectors enable an efficient mechanism for processing multiple subtasks in parallel and focusing on the dominant subtask (The task assigned to the detection branch where the renormalized connection is inserted is the dominant task.).

Facing the practical application requirements of remote sensing satellite videos, we have specially designed a purely tiny object detection problem with the following challenges: 1) Small size: All the objects to be detected in the dataset are extremely small in size. Small objects account for a small proportion of the scene, a small area, and a small number of pixel points. 2) Low image ground resolution: When the image ground resolution is low, the boundaries of small objects are not sharp, and most of the features disappear, making it difficult to distinguish them and resulting in missed detections. 3) High density of distribution: If the objects to be detected have a high distribution density, it is easy to lead to missed detection. 4) Low contrast: Due to the long distance of satellite image shooting, small objects are not rich in color and texture and have low contrast with the background, which can easily lead to detection errors. 5) Interfering objects: Objects with similar top surfaces are scattered throughout the scene and can easily cause confusion. A new tiny object detection dataset, called the IPIU dataset, is being created to address all these challenges. This helps us to construct the task with the highest scale-preferred in this article. In addition, we conducted experiments on four different datasets with different ranges of scale distributions.

We validate the performance of the Renormalized Connections on 17 well-designed detection architectures with three dominant backbone networks as well as feature extractors with different connection layers and densities, on four satellite imagery datasets and a natural small object detection dataset. Extensive experiments demonstrate the effectiveness and high efficiency of the RCs. On a variety of scale-preferred tasks, the $n21$Cs eliminate two aspects of interfering activations and allow the network to fully focus on the target to be detected, significantly improving learning efficiency and detection performance. Since all experiments can be trained and tested on a graphics card with limited video memory, the range of applications is greatly enhanced!

In summary, our contributions are as follows:

(1) We propose a Knowledge Discovery Network on a single-branch detector to implement the renormalization group theory on feature extraction network and find that the feature extractor produces independent multi-level features and these different layers of features act synergistically on the detection task which is consistent with the ``each in its own way'' idea of multi-level FPN. We refer to the mechanism of KDN's RC as ``synergistic focusing''.

(2) Based on the observations derived from KDN, we mathematically abstract a class of renormalized connections with different coefficient ratios called $n21$Cs and generalize them to the FPN-based multi-branch detectors, thereby significantly eliminating negative samples and interfering activations in various scale-preferred tasks. For FPN-based $n21$Cs, they implement both ``divide-and-conquer'' and ``synergistic focusing'' mechanisms as well as appropriate subtask assignment for parallel training of multiple detection branches on scale-preferred tasks. The RCs are customized connection methods.

(3) We create a difficult scale-preferred task on satellite imagery applications, that is IPIU Dataset, a purely tiny object detection dataset, and provide a benchmark to fairly compare the performance of different detection algorithms.

(4) Extensive experimental results, including mAP\&AR, AP and loss curves, saliency maps, visualizations, etc., have verified that the Renormalized Connection approach can renormalize the information flow in the forward and back propagation phases of network learning and adjust the learning process to a more correct direction.

The remainder of this article is organized as follows. Section \ref{related_works} summarizes typical connection modules and state-of-the-art satellite image object detectors. Section \ref{method} introduces the application of renormalization group theory to feature extraction, i.e., the Knowledge Discovery Network. Based on the observations of the network, we construct a new type of connection called a Renormalized Connection (RC). It can be generalized to a variety of detectors. Section \ref{experiments} reports the experimental setup, performance comparisons, visualization results, and discussions on five representative scale-preferred or scale-diversified datasets. Finally, Section \ref{conclusion} concludes the article and identifies future work.

\section{Related Works}
\label{related_works}

In this section, we review representative connection modules in the history of deep networks and survey novel satellite image detection methods. The connection strategies \cite{he2016deep}\cite{vaswani2017attention}\cite{10477580} in network architecture design have made conspicuous performance gains in a variety of computer vision tasks. Research on satellite applications from different perspectives has flourished in the community.

\subsection{Connection Modules in Deep Networks}
\begin{figure}
\centering
\includegraphics[width=3.5in]{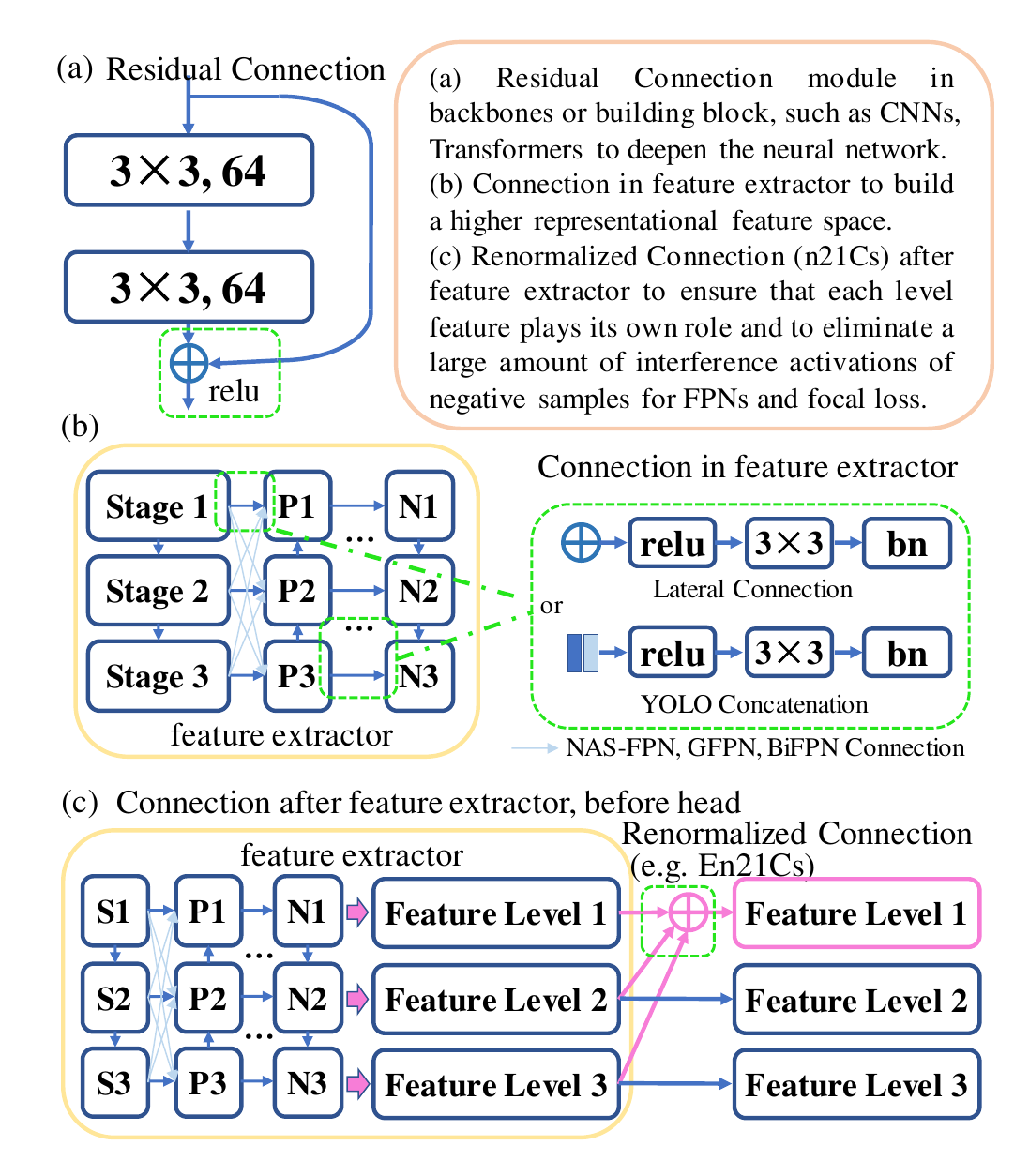}
\caption{Three types of feature connection methods. (a) Residual connection is designed to deepen the network by introducing a non-linear computation after summating the input and building block features, i.e., the ReLU unit. (b) gathers the design structures of different feature extractors with different connection layers and connection densities such as GiraffeDet \cite{jiang2022giraffedet}, RTMDet \cite{lyu2022rtmdet}, NAS-FPN \cite{ghiasi2019fpn}, BiFPN \cite{tan2020efficientdet} and GFPN \cite{xiao2023global}. Lateral connection \cite{fpn} is an independent connection module including summation or concatenation, ReLU layer, $3\times3$ convolution layer, and batch normalization layer. In addition to lateral connections for feature maps of the same size, there are also connections for feature maps of different sizes (indicated by light blue arrows). (c) shows the Renormalized Connection (n21Cs), which is used to renormalize the output of the feature extractor and form a new input to the head net. This is an example of Economic $n21$C, embedded in Feature Level 1 only. It only uses the output features of the designed feature extractor as inputs and does not change the internal connection structure of the feature extractors. Moreover, this kind of Renormalized Connection is a linear connection structure and does not contain additional non-linear operations or any learnable layer (except for the projection operation) and normalization layers yet works well.}
\label{comparison}
\end{figure}

It is mentioned in ResNet \cite{he2016deep} that deep networks cannot automatically learn the structure of identity mapping and need to manually add residual connections. Consequently, in addition to designing the basic module of the backbone network, in the detection task, researchers have also worked on designing feature extractors with different connection methods. As shown in Fig. \ref{comparison}, we draw the Renormalized Connection (as known as $n421$C of the generalized RC) with the residual connection, the connection of multiple FPN-based feature extractors.

Differences between Renormalized Connection and various existing connection methods: 

1) Different scope of action. Residual connection is used to deepen the backbone network, feature extractor, or any building block in a variety of deep networks. Lateral connection is used for feature fusion within the feature extractor. And YOLO concatenation from v3\cite{redmon2018yolov3} to v8\cite{yolov8} is used for feature fusion in each base module. The Renormalized Connection is used for feature renormalization for the feature extractor to provide a ``synergistic focusing'' feature pyramid for the detection head and to eliminate interfering activations for a large variety of scale-preferred tasks. RC is a bridge between the feature extractor and the detection head, directing the multi-level outputs of the feature extractor to be correctly used for the feature supply of the embedded branch in the detection network. The connection strength, embedding position (one or more detection branches), and the number of RCs can be flexibly set to meet different task requirements.

2) Different purposes. The residual connection and lateral connection methods are used to increase the capacity of the model. In contrast, the Renormalized Connection is used to renormalize the information flow at all stages of network learning (rearranging the feature flow in the forward propagation phase and reorienting the gradient flow in the backpropagation phase), adjusting the learning process in a more correct direction.

3) The complexity of the connections is different. The residual connection and lateral connection work well with nonlinear operations involving nonlinear activation units, batch normalization layers, or new convolutional layers. These non-linear connection methods can be considered as a completely new module. The RC produces a very good renormalizing effect using only linear connections. However, adding nonlinearities to the linear RC, such as the $1\times1$ conv and the normalization layer, can significantly reduce performance.

The difference between the RC and the bottom-up path in PAFPN \cite{liu2018path} is that the inputs to the economical RC (E421C, only for the small object detection branch corresponding to the $P_3$ level in FPN) are the three direct outputs of the top-down path in the FPN, whereas the bottom-up path in PAFPN uses only one of the features of the top-down path in the FPN ($P_3$) as the input to the small object detection branch. Moreover, PAFPN acts as a feature extractor and the RC can be used on top of it as a bridge to connect the feature extractor and head. 

The Renormalized Connection has a different scope of application and functionality compared to the RoI extractors\cite{liu2018path}\cite{rossi2021novel} (feature fusion methods for RoIs). Since only the two-stage object detector contains a RoI pooling/RoI align layer, the RoI feature fusion methods \cite{liu2018path}\cite{rossi2021novel} can only be used on two-stage detectors. In contrast, the RCs can be used with a RoI extractor on two-stage detectors and greatly eliminate the inference activations of one-stage detectors. The RoI extractors are all designed for scale-diversified datasets, fusing only features within the RoI and only on two-stage detectors. The RC is a connection between the feature extractor and head to renormalize information flow for various detectors.

On top of the broader and more powerful detection architectures built by various connection methods, RCs can further improve the accuracy of scale-preferred detection tasks. Most importantly, the simplest RC, i.e., E421C, can make the above connection methods work well in a variety of scale-preferred tasks, thus greatly extending their range of applications.

RCs achieve the design idea that all the detection branches play their own roles without interfering with each other, focusing on the dominant task. RC inherits the ``divide-and-conquer'' and multi-subtask parallel learning mechanisms of FPN-based detectors, and brings a ``synergistic focusing'' mechanism to detectors to solve scale-preferred tasks. 

\begin{figure*}[!t]
\centering
\includegraphics[width=7.16in]{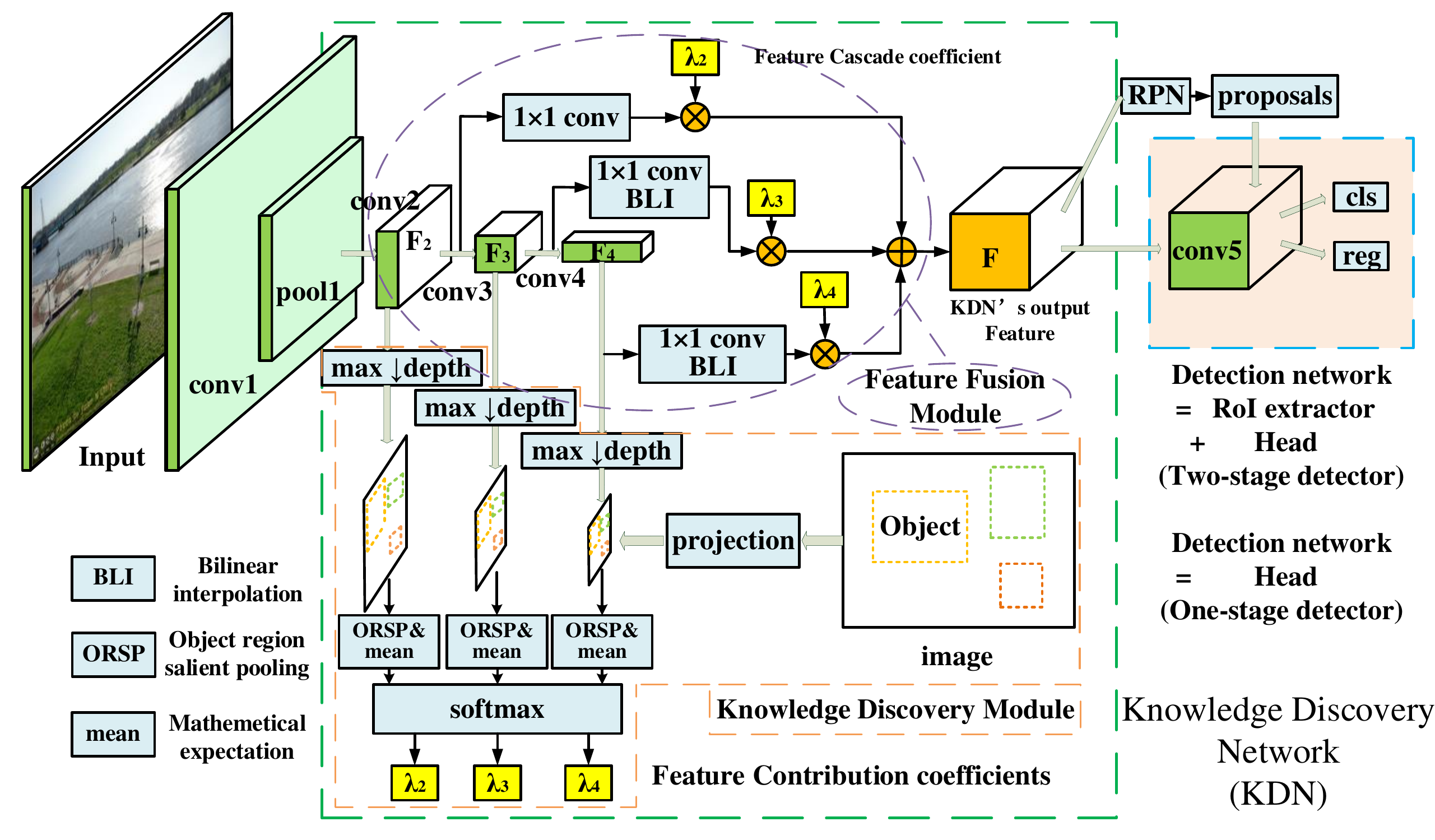}
\caption{The framework of the Knowledge Discovery Network (KDN).} 
\label{kdn_diagram}
\end{figure*}

\subsection{Satellite Imagery Object Detection}
For small object detection, Cao et al. \cite{10364853} used physical simulation images as prior knowledge to extract the target's key geometric features through a dynamic matching method. Yin et al. \cite{10379602} proposed a lightweight hybrid attention block for ship detection. Wang et al. \cite{9208789} introduced a random projection feature for characterizing blocks into feature vectors to solve the problems with dynamic background and scale variation targets.
Yuan et al. \cite{10328639} proposed a classifier for vehicle classification under foggy weather conditions. Yuan et al. \cite{9584924} captured context to enhance classification scores by enlarging receptive field regions and designing head networks. Zhang et al. \cite{10495059} used super-resolution and cross-modality information fusion.

Moreover, recently, satellite remote sensing images have plenty of applications, such as cloud detection \cite{10311415}\cite{10323364}, multiclass object counting and localization \cite{10381622} etc.

\section{Methodology}
\label{method}
In this section, we present methods to cope with the scale-preferred tasks. First, we introduce renormalization group theory for feature extraction. Specifically, we construct a renormalized feature distribution for various feature extractors and redirect the learning for the detector. We implement the renormalization group theory on comprehensive feature extraction by designing a Knowledge Discovery Network (KDN). We then use the observations of the KDN to build a new type of connection, called a Renormalized Connection, denoted as $n21$C ($n \in R$). This $n21$C method can be generalized to a variety of detectors. The simplest and most generalizable type of renormalized connection is the Economical 421 Connection. Except for the 421C, we have constructed the 521C and the complete form of RC to depict the RCs from a broad perspective. Now, let us take a look at the promising renormalization methods.

\subsection{Renormalization Group Method for Feature Extraction}
\label{renorm}
A common feature of problems where renormalization appears is the presence of infinitely many particles (real or virtual) in the interaction between them, often with an emergent behavior that can be very different at different energy scales \cite{mastropietro2023renormalizationgeneraltheory}. For example, we use a tree to illustrate. The renormalization method focuses on the self-similarity of the tree since each branch of the tree is similar to the traits of the whole tree, the branches are similar to smaller branches, and so on. It is the features hidden in these iterative operations that the renormalization group \cite{kadanoff1966scaling} wants to extract, hoping to unearth the relationships between the system at different scales.

On a deep neural network, as the network moves from lower to higher layers, the features it extracts become more and more abstract and involved in the nature of the ``whole''. At a deeper level, deep neural networks are not simply a stack of neural networks, but can actually find some hidden symmetries and constraints in the data. This means that the features extracted by deep neural networks will be ``slow variables'' that are not affected by noise, which is consistent with the idea of renormalization, that is, mutually independent variables with appropriate degrees of freedom. 

The basic physical idea underlying the renormalization group approach is that the many length or energy scales are locally coupled \cite{wilson1975renormalization}. Inspired by the methodology of the renormalization group in statistical mechanics, we build a renormalization method for comprehensive feature extraction. The aforementioned idea has resulted that there is a cascade effect on the whole system. In multi-scale feature extractors, the different but contiguous feature scales also have a cascade effect. For example, the behavior of the feature stage ($C_3$ in Faster R-CNN) is responsible for detecting tasks and is directly sent to the head network are assumed to be primarily affected by the nearby feature stages, e.g., the $C_2$ and $C_4$. The single-branch detector uses only single-scale features (one stage) for the detection task, and the remaining feature stages constitute the feature extractor of the detector. To verify that there is a cascade effect in the feature extractor, the single-branch detector is the best choice. Once the base model is selected, we build the cascade for the renormalization method.

There are two principal properties (\cite{wilson1975renormalization} used features) of the feature cascade. The first property is scaling. The behavior of feature extraction for intermediate feature stages tends to be identical except for a change of scale. We will verify it in Section \ref{insights}. We focus here on describing the second property, the existence of amplification and deamplification as the cascade develops. That is, we have to determine the quantitative relationship and interaction between the contiguous feature cascade. Corresponding to a single-branch detector, we have to determine the extent to which the previous and subsequent feature stages affect the detection task. The amplification effect implies that there are complementary and synergistic effects between any successive two feature scales. The deamplification effect, on the other hand, indicates a mutual exclusivity and correlation between these features. 

Next, we perform the renormalization group analysis of the detection system to understand the relationship between contiguous feature stages in the feature extractor.

\subsection{Knowledge Discovery Network --- the Implementation of Renormalization Group Theory in Feature Extraction}
\label{kdn_implement}
We design a Knowledge Discovery Network (KDN) to implement renormalization group theory in feature extraction. The purpose of using renormalization group theory is to renormalize the output feature flow of the feature extractor to further guide the detector to find a suitable learning direction quickly. As mentioned above, the single-branch detector is the best choice to verify the cascade effect in the feature extractor. Therefore, we construct the KDN on a single-branch detector as shown in Fig. \ref{kdn_diagram}.

The first step in renormalization is to determine the feature cascade consisting of features at different scales in the feature extractor. Each scale of features in the cascade represents an independent variable. The independent degrees of freedom for feature extraction in single-branch detectors is $3$. The renormalization group finds the relevant variables, and once we have combined all the relevant variables, we will have ``a view of the mountains''. The combined relevant variables are used to describe the features that govern the more comprehensive behavior of the system and are independent of noise and local fluctuations.

In the second step, we should calculate the amplification and deamplification factors $\lambda_i$ for each scale ($i=2,3,4$). This consists of an explicit statistical averaging over the knowledge we have extracted from each information source in the feature cascade. The third step is to construct a renormalization group for the detection head. 

For example, we use Faster R-CNN as a single-branch detector. The backbone provides three stage features to compose a feature extraction network. Thus, the feature cascade consists of ${F_2, F_3, F_4}$ with a scale multiplier of 2. The amplification factor or deamplification factor is generated by mining the knowledge in the information source. For our task, the information sources are the objects in the dataset. The KDN translates the knowledge into an amplification or deamplification factor for each stage in the feature cascade. To obtain knowledge from all information sources, we need to traverse all the images in the dataset. Naturally, traversal can be achieved by training the network. After one epoch of training, all images are traversed once.

\subsubsection{The Architecture of KDN}
\label{arch_kdn}
KDN is located in the middle part between the feature extractor and the detection network.\footnote{The detection network in this work specifies a head network with one or more branches rather than an entire detector, where each branch comprises a classification subnetwork and a regression subnetwork.} The KDN consists of two parts: a Knowledge Discovery Module and a Feature Fusion Module. The Knowledge Discovery Module executes the first and second steps of the renormalization method. The Feature Fusion Module generates the renormalization group for the detection network. After training, we can get the final factor set for inference, which determines the amplification or deamplification factor for each independent variable in the feature cascade.

\subsubsection{Knowledge Discovery Module}
\label{kdm}
The Knowledge Discovery Module (KDM) extracts information from each object, then generates useful knowledge through statistical analysis, and finally produces the amplification or deamplification factor for each scale in the feature cascade. Because we only focus on small objects, after a series of down-sampling operations, the remaining feature maps have a smaller width and height; the effective feature region of small objects also remains smaller. For example, a $4\times4$ pixel object in the original image is left with 1 effective pixel in the output feature map after two downsampling layers with stride 2. Therefore, we select the most salient feature value in the projection region of each feature scale, which is sufficient to represent the information of the object on the feature map.

First, the object's position is projected onto the feature map at each level. Then, an effective feature region of the object is determined for each scale in the feature cascade. The salient information of an object can be extracted from its effective feature region. Given an input image and a feature scale, the KDM extracts the salient information of each small object (on the projected feature region) and stores it in a stack. After collecting information about all objects of interest in the image, the KDM calculates the mathematical expectation of the salient stack to represent the global knowledge of all objects in the image mapped to that feature level, and from this determines an amplification factor or deamplification factor for each scale. If the factor value is greater than 1, it is determined as an amplification factor, otherwise, it is determined as a deamplification factor. The amplification factor corresponds to a relevant variable. A variable whose effect is deamplified is called an ``irrelevant variable''. The number of relevant variables is equal to the number of amplification factors $\lambda_i$ greater than one.\footnote{In this article, we use a looser $\lambda_i \geqslant 1$ condition.} Finally, the relevant variables are combined to characterize the features that govern the more comprehensive behavior of the system for detection tasks.

For each image (has $n$ objects) and its corresponding feature cascade (with degrees of freedom as 3), KDM performs the same calculation as follows. 

Step 1: Reduce\_max operation (returns the maximum value of each channel in the feature maps):
\begin{equation}
rm(F_l), (l=2,3,4)
\end{equation}

Step 2: Object region projection:
\begin{equation}
orp_i(rm(F_l)), (i=1,\ldots,n; l=2,3,4)
\end{equation}

Step 3: Object (projected) region salient pooling:
\begin{equation}
\text{ORSP}_i(orp_i(rm(F_l))), (i=1,\ldots,n; l=2,3,4)
\end{equation}

Step 4: Mathematical expectation calculation:
\begin{equation}
\lambda_l^e=\frac{\sum_{i}{\text{ORSP}_i(orp_i(rm(F_l)))}}{n}, (i=1,\ldots,n; l=2,3,4)
\end{equation}

Step 5: Softmax operation:
\begin{equation}
\lambda_l=\frac{exp(\lambda_l^e)}{\sum_{j}{exp(\lambda_j^e)}}, (j,l=2,3,4)
\end{equation}

Step 6: If $\lambda_i \geqslant 1$, $(i=2,3,4)$: returns an amplification factor; else: returns a deamplification factor. 

Step 7: The amplification factor corresponds to a relevant variable in the feature cascade $\{F_l, l=2,3,4\}$. Retaining all the relevant variables in the feature cascade as $\{F_i, i=np.where(\lambda_i \geqslant 1)\}$ \footnote{np.where() returns the indices of the elements that are non-zero.}.

\subsubsection{Feature Fusion Module}
\label{ffm}
As usual, the feature extractor has different ``level'' features of different scales and dimensions. The content of different ``level'' features is also different; some of them focus on representing details, and others focus on representing more abstract features. To capture comprehensive features of an object, it is beneficial to integrate different levels of features. The Feature Fusion Module (FFM) aims to generate a more complete feature description for the detection network by integrating the different ``levels'' of features that make up the feature cascade.

The input for each branch in the detection head has a fixed scale and dimension. Consequently, the KDN has to output features of fixed size. Features of different ``levels'' must be mapped to the same size and dimension. For example, if the detection network requires the input of features from the middle level, FFM should map the high-level features to the middle level and low-level features to the middle level to ensure that features from all levels have the same size and dimension. Unlike the form of feature fusion that connects the layers within the feature extractor - direct summation or concatenation operations - we wish to fuse the features by means of a corresponding amplification or deamplification factor. The amplification or deamplification factor is the extent to which each ``level'' of features in the feature cascade contributes to the overall feature. After obtaining features of the same size and dimension in FFM, we use the amplification or deamplification factors generated by the Knowledge Discovery Module to output a comprehensive feature for detection as,

\begin{equation}
F={p(F_2)}\times{\lambda_2}+ {BLI(p(F_3))}\times{\lambda_3} + {BLI(p(F_4))}\times{\lambda_4},
\label{ffm_eq} 
\end{equation}
where $p()$ is the $1\times1$ conv projection.

The Feature Fusion Module generates the comprehensive features in Step 8 after Step 7 of the KDM.

Step 8: During training, the combination of all independent variables multiplied by the corresponding factors yields the comprehensive features as described in Eq. (\ref{ffm_eq}).

\subsubsection{The Final Factor Set}
\label{final_factor_set}
During the training process, for each training batch, we have access to the amplification or deamplification factor for each independent variable in the feature cascade. After traversing the entire dataset, we obtained 3 stacks containing the factor sets for all the images, where the 3 factors correspond to the independent variables of the three degrees of freedom in the feature cascade. To generalize to inference process, we have to store the factor set of each independent variable as a stack and compute the mathematical expectation of each independent variable. The set of factors used for inference is defined as

\begin{equation}
\lambda_l^\text{infer} = E(\text{stack}(\lambda_{li})),(i=1,2,\cdots,m_l;l=2,3,4)
\end{equation}
where $\lambda_l^\text{infer}$ is the $l^{th}$ factor corresponding to the $l^{th}$ independent variable for inference; $\lambda_{li}$ is the $i^{th}$ factor of the $l^{th}$ independent variable in the training process; $\text{stack}(\cdot)$ denotes the factor stack corresponding to the independent variable; $E(\cdot)$ is the mathematical expectation operation; and $m_l$ is the length of the $l^{th}$ ($l=2,3,4$) factor stack corresponding to the independent variable.

At this point, we can determine whether the factor of each independent variable is an amplification or deamplification factor, and retain all relevant variables corresponding to the amplification factor ($\lambda_i \geqslant 1$).

\subsubsection{The Comprehensive Features of the Inference Stage}
\label{infer_feature}
With the amplification factors and relevant variables in place, the next step is to generate the comprehensive features used for inference.

Step 9: Merging all the relevant variables (corresponds to $\lambda_i \geqslant 1$) outputs comprehensive features for the detection network in the inference stage. The comprehensive feature can be expressed as

\begin{equation}
F^\text{infer}=\sum_{i}{{p(F_i)}\lambda_i\mathbbm{1}(\lambda_i \geqslant 1)},
\label{infer_eq} 
\end{equation}
where $p()$ is the $1\times1$ conv projection and bilinear interpolation operation; $\mathbbm{I}(\cdot)=1$ if $\lambda_i \geqslant 1$, otherwise $\mathbbm{I}(\cdot)=0$.

\subsubsection{The Insights of KDN}
\label{insights}
As shown in Fig. \ref{kdn_diagram}, the KDN inputs three \footnote{It depends on the independent degrees of freedom in the feature cascade.} ``levels'' of features from the feature extractor and outputs single-scale comprehensive features for the detection network. The three scale features with the corresponding amplification or deamplification factors determine the final feature output. In the inference procedure, we found that the KDN generates three equal amplification factors for different types of small object detection tasks. That is, the KDN outputs a set of amplification factors for the feature cascade.

This phenomenon means that although in the initial Faster R-CNN, except for the $C_3$ stage in the feature extractor, the other stages are not directly involved in the detection task, each stage produces a different level of features that are both independent of and complementary to each other. That is, we pre-constructed a set of independent variables in the feature extractor can implement the renormalization group method. Finally, these variables are verified to be independent and complementary to each other. In other words, each independent variable in the feature cascade acts synergistically on the objective detection task without overlap or interference. This observation for the single-branch detector is consistent with the idea of ``each in its own way'' in the multi-branch detectors' feature extractor represented by the FPN. We conclude that each deep network used as a feature extractor produces multilevel features, which certainly have different functions and can act synergistically in the detection task. 

This insight satisfies the first principle feature ``scaling'' of the feature cascade mentioned in Section \ref{renorm}. For example, averaging the $F_{L_1}$ information of an object produces a factor that is very similar to the factor produced by averaging the $F_{L_2}$ information of the object. This further validates our successful application of the renormalization group theory to the comprehensive feature extraction problem.

\subsection{The Renormalized Connection Methods}
\label{rec}
In the following, we will discuss the renormalized connection methods. Firstly, we will introduce the formulation of the renormalized connection method and then generalize it to a wider range of detection networks, such as multi-branch detection networks using FPN series as feature extractors.
\subsubsection{The Formulation of Renormalized Connections}
\label{formulation}
We formulate the renormalization group method for feature extraction as a renormalized connection. It is well known that any feature in a feature space can be represented by a set of feature bases and their corresponding coefficients, i.e., $f=\sum_i{c_ib_i}$, where $c_i$ denotes the coefficient of the $i^{th}$ feature base; $b_i$ denotes the $i^{th}$ feature base.
By convention, in a feature pyramid extractor, the different feature levels used for object detection are in different scale ranges but share weights in the head. We observed through KDN experiments with a single-branch detector that each of the pre-constructed independent variables in the feature cascade is indeed independent and complementary to each other. 

\begin{figure}
\centering
\includegraphics[width=3.5in]{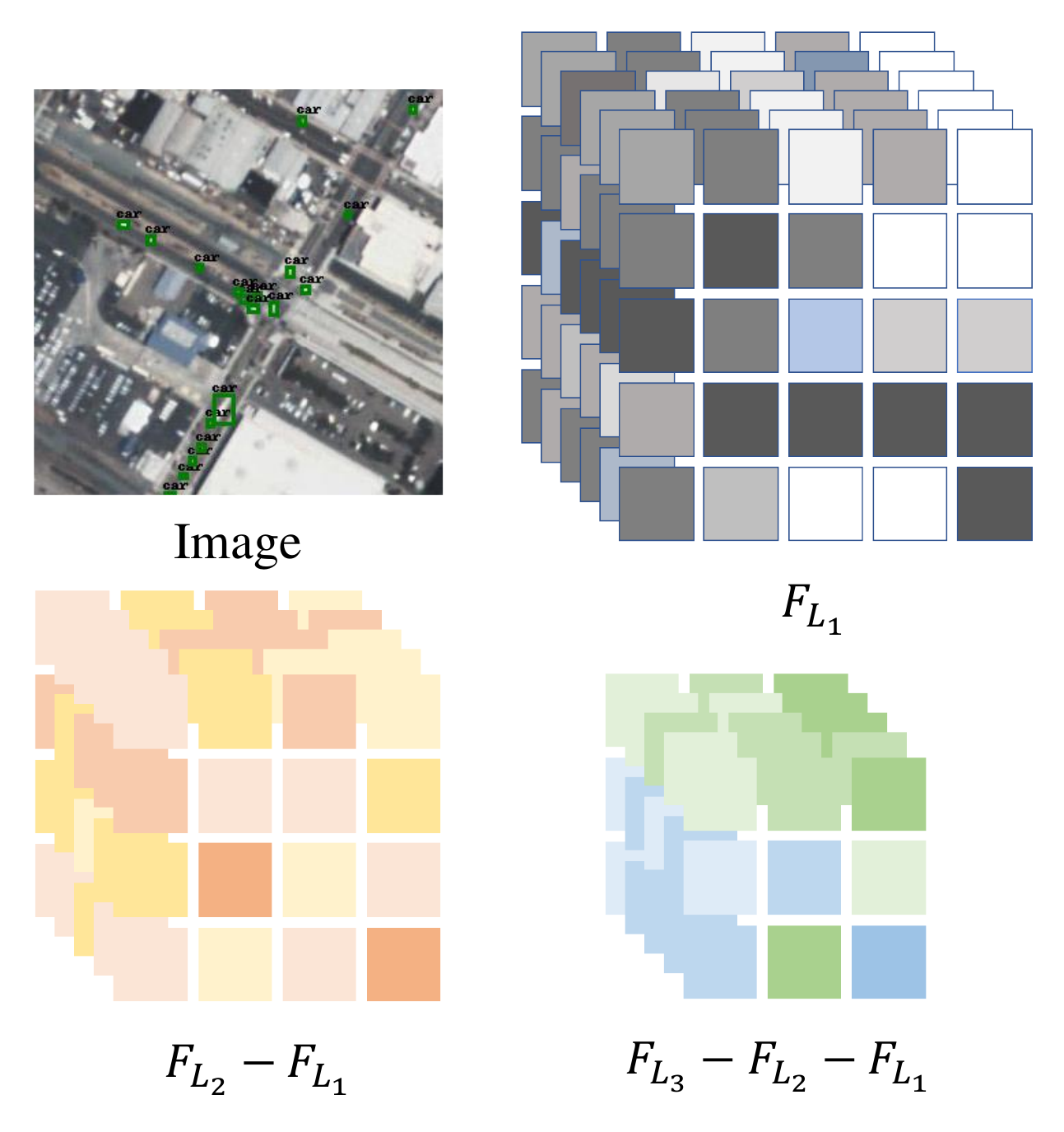}
\caption{A set of independent feature bases in a given feature space.}
\label{feature_bases}
\end{figure}

To ensure independence, we need a set of stricter bases to replace the initial independent variables in the feature cascade to cover the general conditions of feature pyramid extractors. To construct a set of independent bases, we first need to remove the possible correlation part between every two variables in the feature cascade. We formulate a feature cascade with 3 degrees of freedom as $F_{L_1},F_{L_2},F_{L_3}$. The $F_{L_1}$ acts as the first independent variable and maintains the original variable type as $F_{L_1}$. Next, we take the second variable $F_{L_2}$ minus the first independent variable's remainder ($F_{L_2}-F_{L_1}$) as the second independent variable. Finally, we keep the third variable $F_{L_3}$ minus the first and second parts ($F_{L_3}-F_{L_2}-F_{L_1}$) as the third independent variable. Up to this point, three tighter bases of independent variables are constructed, namely $F_{L_1}$, $F_{L_2}-F_{L_1}$, and $F_{L_3}-F_{L_2}-F_{L_1}$, as illustrated in Fig. \ref{feature_bases}.

The renormalized connection is then established according to the conclusions of the renormalization group method of feature extraction. In the case of single-branch detectors, we have a set of uniform amplification factors to form a comprehensive feature, i.e. $F=F_{L_1}+F_{L_2}+F_{L_3}$. We can use the Gaussian elimination method to calculate the connection strength of the set of uniform amplification factors. After that, we get $F_{L_1}+F_{L_2}+F_{L_3}=4\cdot{F_{L_1}}+2\cdot{(F_{L_2}-F_{L_1})}+(F_{L_3}-F_{L_2}-F_{L_1})$. This connection strength ratio is $4:2:1$, and we can generate one kind of renormalized connection called the 421 Connection, or 421C for short. 

In this work, our focus is on difficult scale-preferred tasks, mainly the detection of small objects. While there are many possible sets of factors to construct a renormalized group for feature extraction, we focus on adjusting the amplification factor of the first independent feature to renormalize the feature distribution of the feature extractor. The most direct connection strength comes from the observations of the KDN, forming a 421C connection. The calculation of the renormalized connection strengths for different amplification factors is straightforward, and it is sufficient to follow the same procedure that we used for the calculation of 421C. 

Using 421 as a reference, if we have a deamplification factor of the first irrelevant variable $F_{L_1}$, the first connection strength $4$ will be reduced. Conversely, if we have a larger amplification factor for the first relevant variable, the first connection strength $4$ will be amplified. Under this condition, we can obtain a $n21$C, where $n > 4$ if the first independent variable corresponds to an amplification factor larger than the others; and $n < 4$ if the first irrelevant variable has a deamplification factor. 

The simplest and most specific connection strengths are $\{4,2,1\}$. We use a random search method to find the appropriate connection strengths for each scale-preferred task. And the $n$ in the $n21$C can be changed to adapt to different types of detection architectures. While there are multiple possibilities for connection strengths, as the experiments show, most design parameters are not particularly sensitive to exact values. Consistent with the scaling property of renormalization group theory in the feature cascade, we primarily use in this work the 421C that most typically satisfies this property. Next, we present the generalized RCs of multi-branch detectors based on feature pyramid networks.

\begin{figure*}
\centering
\includegraphics[width=7.16in]{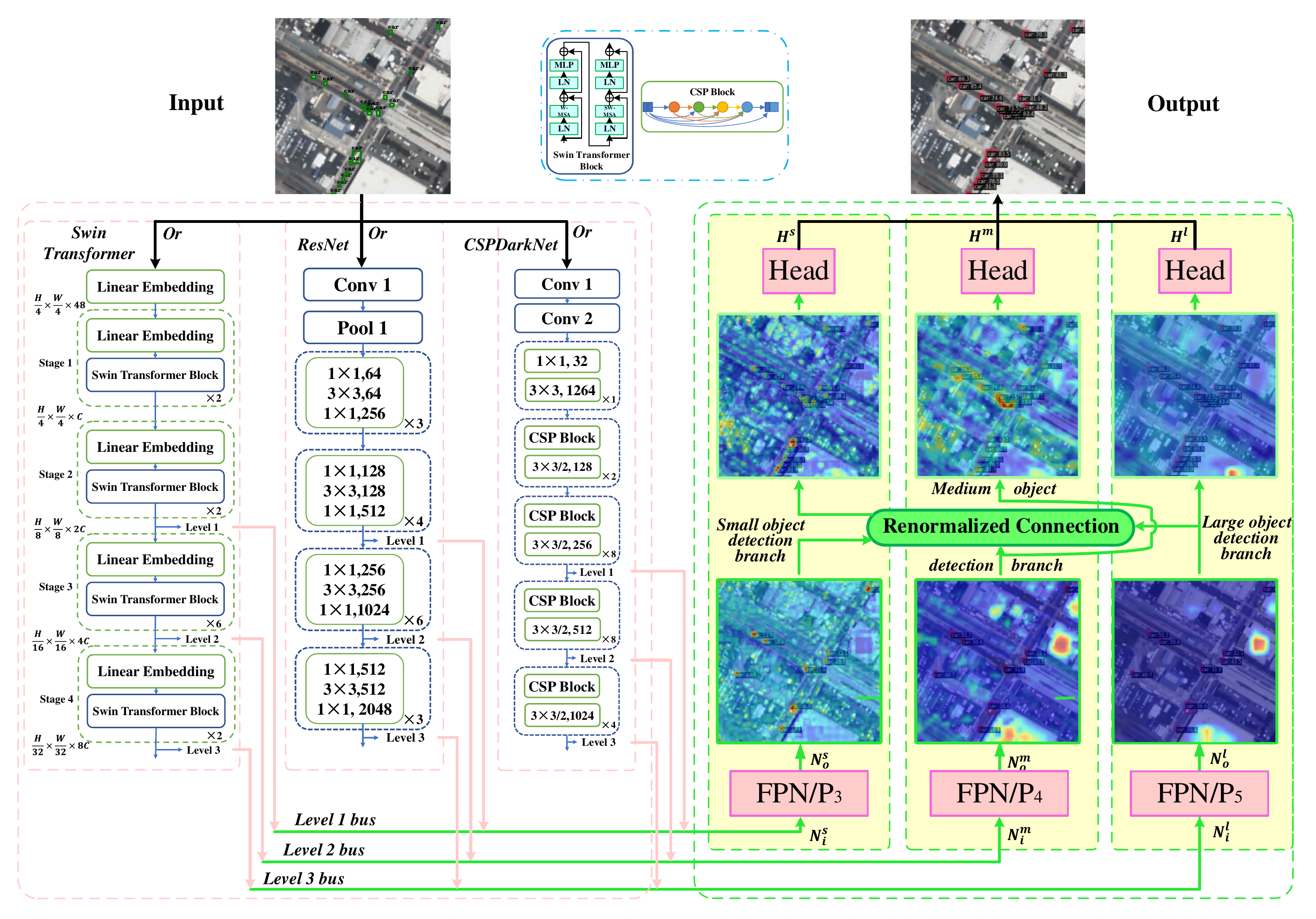}
\caption{The architecture of the detectors embedded with the Renormalized Connection (economical form). Note that the feature pyramid networks have different connection layers and densities as shown in Fig. \ref{comparison}(b). For simplicity, the fine structure of the feature extractor is not shown. Level 4 of Swin Transformer and ResNet are not shown for simplicity.}
\label{framework}
\end{figure*}

\subsection{The Insights of Renormalized Connections}
\label{insight_rc}
When the FPN-based multi-branch detector is used directly to solve the scale-preferred problem, the detection branch with fewer positive samples generates a large number of interfering negative samples and background noise, which cannot be eliminated by using the focal loss function as shown in Fig. \ref{heatmap_comparison}. These negative samples cannot be eliminated by using the focal loss function, because they are necessary samples and information generated during the training process according to the mechanism of ``divide-and-conquer'' and multi-subtask parallel learning of different levels of features of the FPN, and they will only be generated continuously and will not be eliminated. However, in the case of scale-preferred, they seriously interfere with the training process and prevent the network from learning in the right direction.

Therefore, we designed the Renormalized Connection method to implement the mechanisms where multi-scale/level features act synergistically and do not interfere with each other \footnote{including the idea of ``each in its own way'', where each level feature addresses only the sub-task for which it is responsible, without interfering with each other.}, and focus on the dominant learning objective \footnote{Without generating a large number of negative samples from objects at other scales and background noise both interfering activations is equivalent to the design idea of fewer positive sample branches acting in concert with positive sample branches, focusing on the detection objective of the positive sample branches.}. 

On a single-branch detector, since there is only one detection branch, its mechanism for solving the main learning objective is naturally reached, and it is easy to realize the design idea of multiscale features acting in concert and focusing on the total learning objective using RCs. 

However, on the multi-branch detector, multiple detection branches are used for solving subtasks with different learning objectives. It is necessary to set up a pathway between the multi-level feature extractor and the multiple detection branches to guide the input features of multiple detection branches. It is no longer confined to the one-to-one correspondence between each detection branch and each level of features, so as to make the multiple detection branches truly realize the idea of ``each in its own way'' without interfering with each other, i.e., the ``divide-and-conquer'' design idea of FPN. Moreover, due to the introduction of RCs, the design idea of multi-scale features acting together and focusing on the main learning objective is also inherited.

\subsection{Generalized Applications of Renormalized Connections}
\label{generalization_rec}
In the KDN experiments, the effectiveness of the Renormalized Connection was verified on single-branch detectors. In this section, we extend the RCs to multi-branch detectors using the FPN family of feature extractors to test their validity. Sections \ref{s421c} to \ref{f421c} describe methods for applying RCs to both single-branch and multi-branch detectors, where the incorporation of RCs in multi-detection-branch architectures can be subdivided into two approaches: economical and complete. Section \ref{grad} calculates the gradient of the RCs during backpropagation, proving the validity of the $n21$Cs.

\subsubsection{Overview}
\label{sls}
For the more difficult tasks of scale-preferred detection, tiny object detection, and small object detection in satellite imagery, we insert the Renormalized Connections into 17 powerful mainstream detectors with different connection layers and different connection densities in the architecture. The process of applying RC is independent of the backbone, feature extractor, and head structure. Furthermore, to validate the performance of RCs on different scale-distributed tasks, we conducted extensive experiments on four typical satellite imager detection datasets with varying scale preferences, categories of interest and ground resolution. The 17 representative detection or segmentation frameworks we use can be classified into two categories according to the number of branches in their detection networks, single-branch detectors and multi-branch detectors. In the following, we describe the application of the RC to each of these two classes of detectors.

\subsubsection{Applications on Single-branch Detectors}
\label{s421c}
The RCs can be used in single-branch detectors such as Faster R-CNN and R-fcn, where the backbone network is the feature extractor. As shown in Fig. \ref{kdn_diagram}, KDN is a typical example of RC in a single-branch detector. This type of feature extractor can construct a set of independent variables as the feature cascade with 3 or 4 degrees of freedom, depending on the number of stages in the feature extractor. Since single-branch detectors have only one detection branch in the detection head, the RC is naturally inserted into the unique detection branch. In this article, we connect all the feature scales in the feature extractor to output comprehensive features of the whole image for the detection network. The RoI pooling layer in the detection network then extracts features from each object and feeds them into the head network for classification and regression tasks.

\begin{figure}
\centering
\includegraphics[width=3.5in]{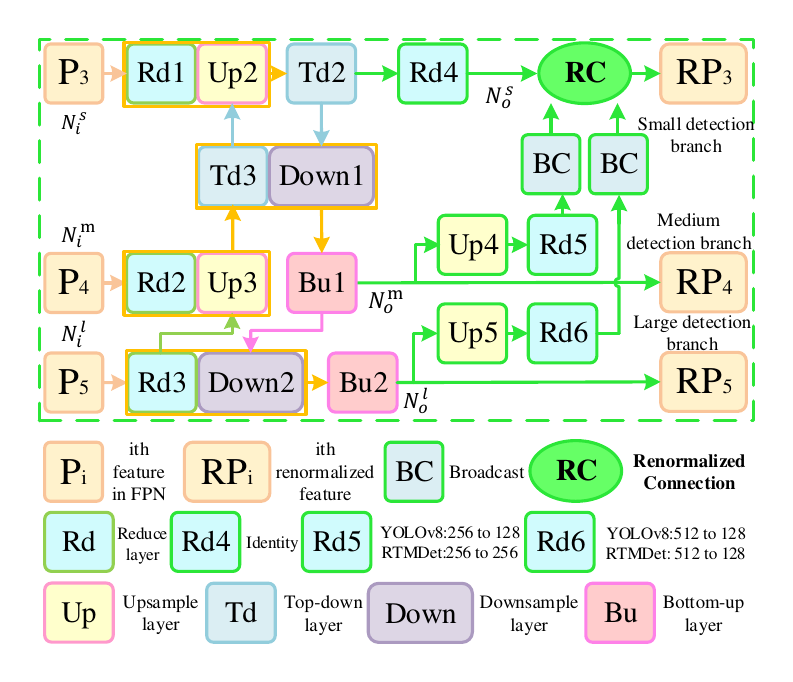}
\caption{Example structure of an economical Renormalized Connection in FPN-based multi-branch detectors.}
\label{rc}
\end{figure}

\subsubsection{Economical Applications on Multi-branch Detectors} 
\label{e421C}
Since the FPN series of feature extractors have multi-level features, there are various forms of constructing RCs on them. Depending on the task to determine on which or which scales of features in the feature extractor to insert the RC, we devised one of the simplest approaches that focus only on the first branch, the small object detection branch, as shown in Fig. \ref{framework}. This simplest and most economical form uses only the first three levels of features $\{P_3, P_4, P_5\}$ as the feature cascade, in which each feature scale represents an independent variable. These independent variables with 3 degrees of freedom are fed into the RC operator. The RC outputs comprehensive features for the small object detection branch. Although the RC is only inserted into the small object detection branch, all features in the feature cascade are renormalized. That is, medium and large object detection branches are also affected by the Renormalized Connection operation. The right part of Fig. \ref{framework} shows an example of input features (multilevel features of the FPN) and output features (renormalized features) of the economical Renormalized Connection. 

We stabilize the connection strength as $\{4,2,1\}$ which satisfies the scaling invariance property of the feature cascade in the multi-level feature extractor to build a 421C at first. In our experiments, we can see that this simple and specific form of economical RC performs well in a variety of different scale-preferred tasks. Therefore, in this work, we focus on the simplest form of renormalized connection that satisfies the renormalization group theory and refer to this economic connection as E421C. We have equipped the E421C with multifarious multi-branch detectors with different architectures.

In addition, we investigate different connection strengths to construct different types of RCs. To uniformly represent the different connection strengths of Renormalized Connections, we use E$n21$C, where $n \in R$. It is worth noting that the RC does not obey the scaling invariance property of the renormalization group when $n \neq 4$. Therefore, we mainly verify the effectiveness of E421C in our experiments. Other types of connections are also reported by randomly searching for connection strengths. For each Renormalized Connection with different connection strengths, we use a simplified connection strength ratio to determine the $n$ in the E$n21$C. 

Apart from the effective E421C, we also investigated in detail the performance of the $\{5,2,1\}$ connection strength on different scale-preferred tasks. This type of connection is termed E521C. The doubling of the $P_3$ information in the input of E521C compared to E421C suggests that 521C is more focused on the small object detection branch than the 421C. Experiments show that E521C performs better or worse than E421C on different detectors (e.g. YOLOv7 and YOLOv8) and on different scale-preferred tasks. Fig. \ref{framework} and Fig. \ref{rc} illustrate the detection pipeline containing economical Renormalized Connections. The first three levels in the feature extractor are rearranged to form the renormalized features and redirect the optimization during training.

The vanilla economical Renormalized Connection is a sparse linear connection that does not require additional learnable layers, non-linear computations, and normalization operations. It minimizes modifications to the multi-branch detector and ensures high inference speed. The E$n21$C is not only suitable for the most difficult scale-preferred task, i.e., purely tiny object detection task, but also improves the accuracy of other scale-preferred tasks and scale-diversified tasks. In addition, we investigate connection methods that add three other representative deep learning techniques ($1\times 1$ conv, norm layer, activation) on top of the linear E$n21$C. 

The applications in this subsection are based on the renormalization group theory, which was first applied to the tiny object detection task but has been well generalized to various scale-preferred tasks. Our goal is to pursue economy, efficiency, low density, and less computation in satellite image processing. The E421C achieves this goal and extends the application of renormalization group theory to feature renormalization. It can be used not only for difficult scale-preferred problems but also for scale-diversified datasets. We note that E421C is a linear connection method that does not increase the computational overhead of the detectors, yet helps the multi-level feature extractor to generate more independent and correct features, which is especially suitable for difficult scale-preferred tasks. Due to its robustness and generality, we mainly discuss the simplest linear E421C in this work. Designing better backbones or feature extractors as well as solving general object detection problems is not the focus of this work, so we opt for the simple design described above.

\begin{figure*}[!t]
\centering
\includegraphics[width=7.16in]{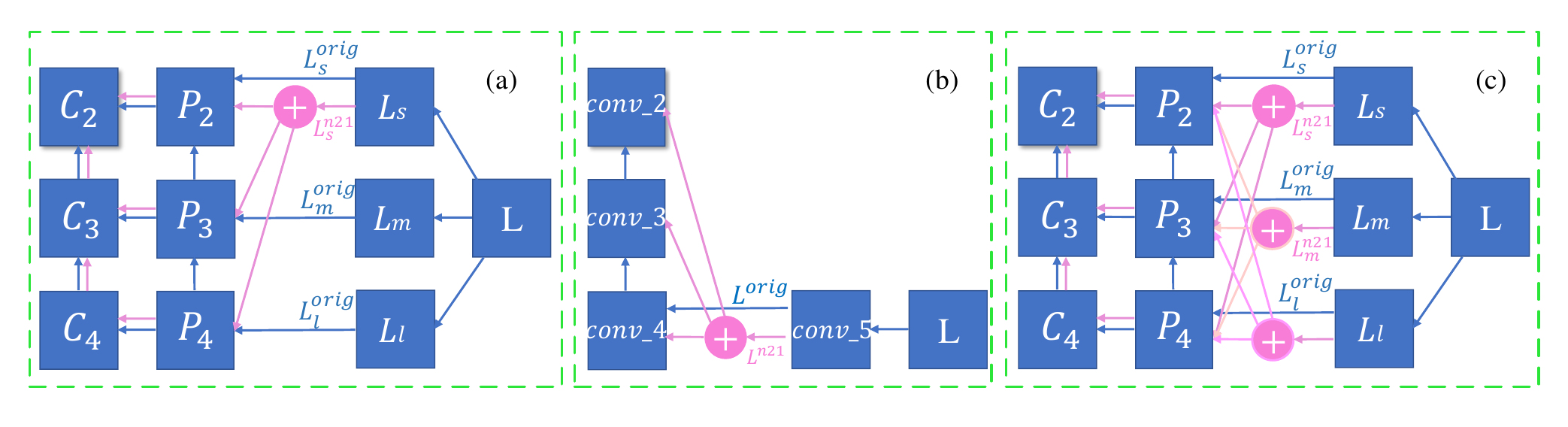}
\caption{The partial backpropagation computational graph of the three forms of Renormalized Connection. (a) Economical $n21$C on multi-branch detectors; (b) Renormalized Connection on single-branch detectors (KDN); (c) Complete $n21$C on multi-branch detectors. }
\label{grad_diag}
\end{figure*}

\subsubsection{Complete Applications on Multi-branch Networks}
\label{f421c}
In the human brain, the density of interneuronal connections, i.e., synapses, is inversely proportional to memory capacity, and the combined effect of the two processes of strengthening important neural connections and degrading unimportant neural connections may lead to continuous optimization of the structure of the human memory system. Inspired by this brain neuroscience research, we have established a complete form of RC, that is, the use of all levels of features in the feature cascade to achieve a renormalized connection before the classification and regression tasks in each detection branch.

To validate the renormalization effect of the complete form, we experimentally compared the complete form of RC with the simplest economical form 421C mentioned in Section \ref{e421C} on the extremely difficult scale-preferred task IPIU pure tiny object detection dataset. We find that the complete connection form reduces the mAP for small objects by a large margin (2.6\%, $26.6\% \rightarrow 24.0\%$) and severely slows down inference ($16.88 \text{ms/f} \rightarrow 22.51 \text{ms/f}$) when compared to the E421C. This result implies that the complete form is incapable of achieving the renormalized feature functions for difficult scale-preferred tasks. Therefore, to follow the mechanisms that strengthen important neural connections in the human brain and to adhere to the principle of Occam's razor, we will focus in this article on an economical approach to applying Renormalized Connections to multi-branch detectors.

The complete form of renormalized connections has more connection strengths than the economical form. Since uniform connection strength does not further improve the performance on the basis of the simplest E421C, we only provide an adaptive feature pooling \cite{liu2018path} inspired connection strength matrix as an alternative complete connection structure. The connection strength matrix is
\begin{equation}
\text{\textit{C}}_{4\times 4}=
\left[{\begin{array}{cccc}
1 & 0.15 & 0.1 & 0.1\\
1 & 0.3 & 0.3 & 0.25\\
1 & 0.25 & 0.25 & 0.25\\
0 & 0.3 & 0.35 & 0.4\\
\end{array}}\right].
\label{eq11}
\end{equation}

\subsection{Gradients of Renormalized Connections}
\label{grad}
The Renormalized Connection renormalizes the information flow in all phases of network learning, rearranging the feature flow in the forward propagation phase and reorienting the gradient flow in the backpropagation phase. The information flow in both phases together determines the learning direction of the network. Next, we analyze the gradient renormalization process in the backpropagation phase.

The essence of the SGD algorithm is that the gradient is large enough that the network stops learning when the gradient disappears. Typically, the gradient value has a very small value of around 0 and can easily disappear with training. Renormalized Connections superimpose 3 or 4 (the independent degrees of freedom of the feature cascade) gradients on each detector branch. We need to consider three cases: (1) a unique detector branch on a single-branch detector; (2) a small object detection branch in multi-branch detectors (economical form); (3) any of the arbitrary detection branches in multi-branch detectors (complete form), as shown in Fig. \ref{grad_diag}.

We know that the deeper layers have larger gradient values due to being closer to the classification layer. As a result, the gradient values of the detection branch output from the Renormalized Connection are larger than those of the original detection branch, which leads to greater parameter tuning, more complete network training, and longer effective training time. Most importantly, RCs can adjust the learning direction of the network to make it learn more efficiently, as we will observe in the experimental results with the AP curve. For example, Fig. \ref{grad_diag} (a) shows the gradient flow of the economical Renormalized Connection, where the blue arrows indicate the original gradient flow and the pink arrows indicate the gradient flow after adding the Renormalized Connection. The renormalized gradient of the small object detection branch can be expressed as

\begin{equation}
Grad(w_s)=Grad_{orig}+Grad_{n21C}
\end{equation}

\begin{equation}
Grad_{orig}={\frac{\partial{L_l}}{\partial{P_4}}}\cdot{\frac{\partial{P_4}}{\partial{C_4}}}\cdot{\frac{\partial{C_4}}{\partial{C_2}}}+{\frac{\partial{L_m}}{\partial{P_3}}}\cdot{\frac{\partial{P_3}}{\partial{C_3}}}\cdot{\frac{\partial{C_3}}{\partial{C_2}}}+{\frac{\partial{L_s}}{\partial{P_2}}}\cdot{\frac{\partial{P_2}}{\partial{C_2}}}
\end{equation}

\begin{equation}
Grad_{n21C}={\lambda_4}{\frac{\partial{L_s}}{\partial{P_4}}}{\frac{\partial{P_4}}{\partial{C_4}}}{\frac{\partial{C_4}}{\partial{C_2}}}+{\lambda_3}{\frac{\partial{L_s}}{\partial{P_3}}}{\frac{\partial{P_3}}{\partial{C_3}}}{\frac{\partial{C_3}}{\partial{C_2}}}+{\lambda_2}{\frac{\partial{L_s}}{\partial{P_2}}}{\frac{\partial{P_2}}{\partial{C_2}}}
\end{equation}

In our experiments, we can see that the test accuracy can continue to be improved by continuing to train the FCOS or Swin Transformer adding with the E421C. The gradient of the KDN is similar to E421C. The gradients of detectors equipped with complete renormalized connections are also easier to derive, and we omit the formulaic descriptions of the latter two forms.

\section{Experiments and Discussion}
\label{experiments}
Renormalized Connections are available in the form of economical connections and complete connections. When used for object detection tasks, RCs can be added to different detection branches according to the task requirements. In order to verify whether RCs can improve the accuracy of different types of scale-preferred tasks, we provide experimental results on five representative datasets in this section.

\subsection{Experimental Settings}
\subsubsection{Datasets}
We conducted experiments on five different types of datasets that represent tasks with a diverse range of scale preferences, namely: 1) IPIU, 2) RSOD Aircraft \cite{long2017accurate}, 3) RSOD \cite{long2017accurate}, 4) NWPU VHR-10 \cite{CHENG2014119}, and 5) MS COCO TOD-80 \cite{f}. The first four satellite image datasets cover the following scenarios: 1) Preferences for different scales, i.e., tiny objects, small objects, or objects of various scales; 2) number of categories ranging from 1 to 10; 3) purely tiny objects or objects of various sizes with large size variations; 4) densely or sparsely distributed objects; 5) used for research or applications; and 6) varying degrees of difficulty in detection. The fifth dataset is a natural scene 80-class small object detection dataset. We primarily use MS COCO TOD to train KDN and also to provide a broader perspective for evaluating Renormalized Connection methods.

\begin{figure}
\centering
\includegraphics[width=3.5in]{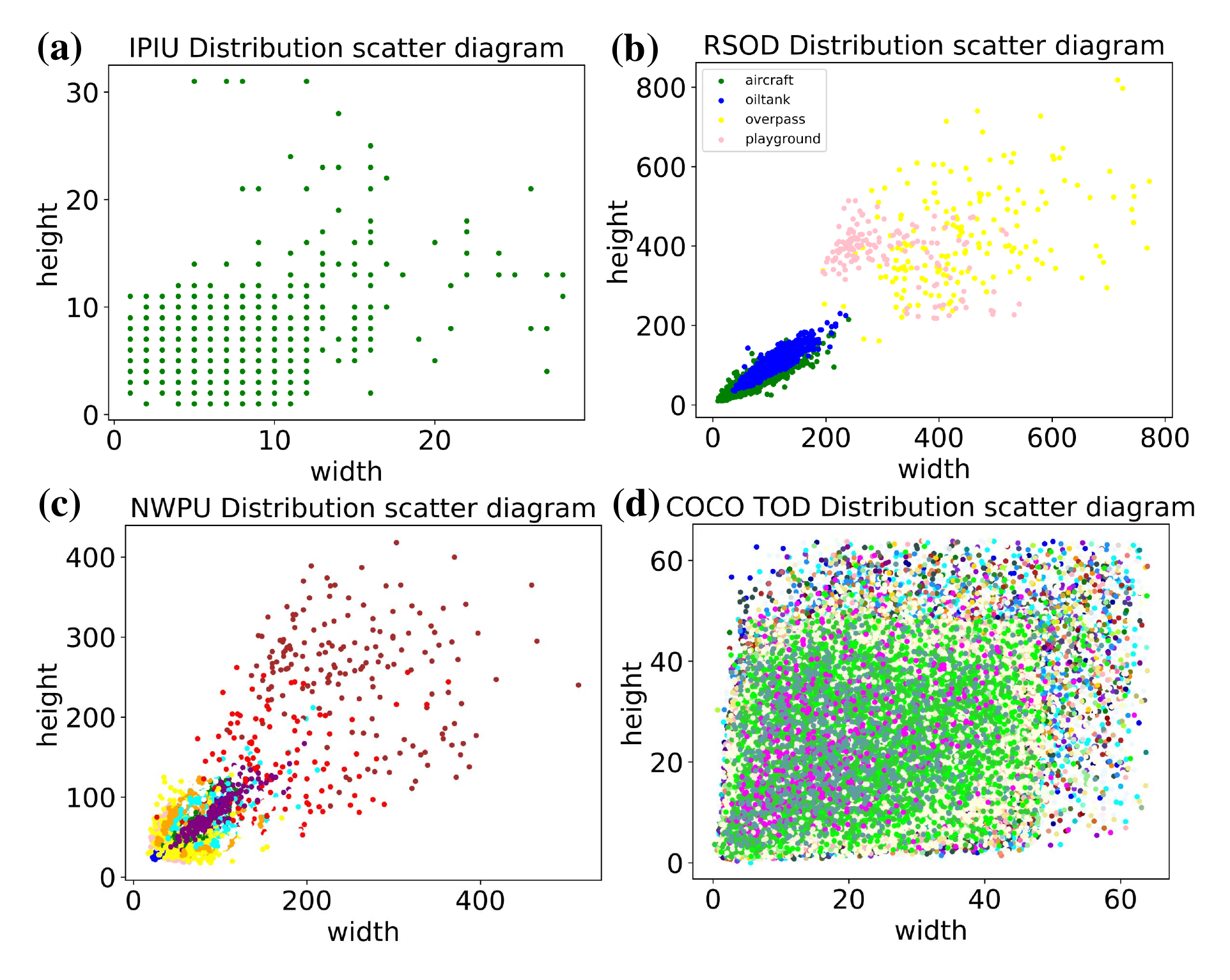}
\caption{Scatter plot of (width, height) points for all instances in the five datasets. (a) IPIU. (b) RSOD and RSOD Aircraft with green points. (c) NWPU VHR-10. (d) MS COCO TOD-80.}
\label{5_dataset_wh_distribution_all_instances}
\end{figure}

\textit{IPIU}: It is a remote sensing video dataset of purely high-density tiny objects. We selected a variety of scenes as detection images and manually annotated all moving vehicles, thus creating a difficult scale-preferred task for this work. The videos were acquired by Jilin-1 HD Dynamic Video Satellite with a ground resolution of $0.91\,m/pixel$ and a frame rate of $10\,fps$. The carefully hand-annotated training and test set were acquired over San Diego Military Harbor, USA, in 2017. We provide the object detection benchmarks in this article. The scenes in this dataset are filled with a wide variety of man-made entities such as roads, overpasses, bridges, ships, containers, buildings, vehicles, cars, plants, airplanes, and so on. The smallest and morphologically diverse entity ``vehicle'' was chosen for this challenge for detection or tracking. 95\% of the objects are between $5\times8$ and $10\times15$ pixels in size. IPIU Dataset is a typical small object detection challenge with all the difficulties mentioned in Section \ref{intro}.

\textit{RSOD Aircraft}: It has densely packed small ``aircraft'' instances with the image resolution of 0.5-2m. It has 153 images containing small objects (whose area is less than ${32}\times{32}$ as defined in \cite{lin2014microsoft}), accounting for 34.3\% of the total 446 images. The total number of instances is 5374 and the number of small instances is 1026 or 19.1\%. In Fig.~\ref{5_dataset_wh_distribution_all_instances} (b) (in green), the size of the instance is densely distributed in (0, 100). This dataset is a suitable representative of the scale-preferred task for small and medium-sized object detection.

\textit{RSOD}: It is a 4-category remote sensing benchmark where the size of the object instances varies a lot. As shown in Fig.\ref{5_dataset_wh_distribution_all_instances} (b), there is a large range of size variations for the 4 categories (aircraft, overpass, oil tank, and playground). The ``overpass'' and ``playground'' instances have a large range of sizes and are much larger than the ``aircraft'' and ``oil tank'' instances. This will visibly divide the 4 categories into two size variation tendencies. The split of the training and validation set is set to $5/1$ as specified in \cite{long2017accurate}. The RSOD dataset is a task with diverse scale distributions.

\textit{NWPU VHR-10}: It is a 10-category remote sensing object detection benchmark. It contains a total of 800 very high-resolution (VHR) remote sensing images. These images were extracted from Google Earth and the Vaihingen dataset and then manually annotated by experts. This dataset represents a task with diverse scale distributions, i.e., a scale-diversified problem.

\textit{MS COCO TOD-80}: MS COCO Tiny Object Dataset is a generic 80-category natural scene benchmark. There are 54,196 images containing 276,702 instances. The training set contains 52,032 images and 265,667 annotated objects. The validation set contains 2164 images and 11,035 objects. There are a large number of varied scenes providing various contexts for 80 types of objects with different shapes. This dataset is used to learn the function of each scale feature in the feature extractor of single-branch detectors. Due to the complexity and diversity of its scenarios and objects of interest, TOD has a small dataset bias \cite{liu2024decades}. Besides, TOD is a typical scale-preferred problem, focusing only on multiple classes of small objects. In summary, this dataset is used to train the KDN and to validate the Renormalized Connection method.

\subsubsection{Implementation Details}
We perform a series of experiments on 21 well-designed detectors and embed the Renormalized Connection in 17 representative baselines for comparison.

For Faster R-CNN, R-fcn, and SSD-FPN, all experiments are performed on the TensorFlow framework and Object Detection API \cite{huang2017speed}. The images are resized to $900\times600$ without any data augmentation. The initial learning rate is set to 0.0003, reduced by a factor of 10 at 900K and 1.2M, and stopped at 1.5M.

For the remaining 18 detectors, all of the experiments are carefully implemented using the Pytorch library based on the MMdetection benchmark \cite{mmdetection}. The input images are resized by a random flip operation. A step-decay learning rate (lr) strategy is adopted with an initial lr of 0.02; the number of linear warm-up steps is 500 with a warm-up ratio of 0.001; the momentum and weight decay are 0.9 and 0.0001, respectively. Schedule\_1x: Training starts with a learning rate of 0.02 (0.002 for YOLOF) reduced by a factor of 10 at the 8th and 11th epochs, and stops at the 12th epoch. Unless otherwise stated, all networks are trained for a total of 12 epochs. Depending on the number of network parameters, each GPU samples $\{1,2,4,8\}$ images per iteration. For YOLOv8: The initial learning rate is set to 0.01. The momentum and the weight decay are set to 0.937 and 0.0005 respectively. And the training is stopped after 500 epochs.

\subsubsection{Evaluation Metrics}
All experiments use COCO-style AP and AR \cite{lin2014microsoft} to evaluate the performance of unseen images. Total inference time and the inference time per frame for all experiments are measured on an NVIDIA GeForce 1080 GPU.

\begin{table*}
\renewcommand{\arraystretch}{1.5}
\setlength\tabcolsep{3pt}
\caption{AP and AR of detectors with RC and baselines on IPIU test set. The images are resized to $600\times600$. YOLOv8 requires $640\times640$ resized images as input and adopts multiscale testing. T: total inference time (s). T/f: inference time per frame (ms). The best AP, AR, and T are in bold.}
\label{table_APAR_ipiu}
\centering
\begin{tabular}{llllllllllll}
\hline
\rowcolor{gray!40}Method & $AP^{0.5:0.95}$ & $AP^{0.5}$ & $AP^{0.75}$ & $AP^{S}$ & $AR_{1}$ & $AR_{10}$ & $AR_{100}$ & $AR_{100}^{S}$ &T(s)${\downarrow}$ &T/f (ms) ${\downarrow}$\\
\hline
\rowcolor{green!40}\textbf{Faster R-CNN-421C (KDN)}&26.3&75.3&9.5&26.3&2.3&18.0&40.1&40.1\\
Cascade R-CNN\cite{cai2019cascade}&13.4&31.5&9.3&13.4&16.5&16.5&16.5&16.5&43s &80.68\\
\rowcolor{green!40}\textbf{Cascade R-CNN-421C}&\textbf{14.3}&\textbf{32.0}&\textbf{10.4}&\textbf{14.3}&\textbf{17.1}&\textbf{17.1}&\textbf{17.1}&\textbf{17.1}&44s& 82.55\\
HTC\cite{chen2019hybrid}& 14.7&35.5&\textbf{9.4}&14.7&18.9&18.9&18.9&18.9&65s &121.95\\
\rowcolor{green!40}\textbf{HTC-421C}& \textbf{15.8}&\textbf{39.7}&9.0&\textbf{15.8}&\textbf{21.0}&\textbf{21.0}&\textbf{21.0}&\textbf{21.0}&73s& 136.96\\

Libra R-CNN\cite{pang2019libra} &11.4&29.2&6.5&11.5&16.7&16.7&16.7&16.7&30s &56.29\\
\rowcolor{green!40}\textbf{Libra R-CNN-421C}&\textbf{12.1}&\textbf{30.4}&\textbf{7.1}&\textbf{12.2}&\textbf{17.2}&\textbf{17.2}&\textbf{17.2}&\textbf{17.2}&31s &58.16\\
FCOS\cite{9229517}& 26.0&73.4&11.0&26.0&38.1&38.1&38.1&38.1&23s& 43.15\\
\rowcolor{green!40}\textbf{FCOS-421C}& \textbf{26.3}&\textbf{73.6}&\textbf{11.3}&\textbf{26.3}&\textbf{38.6}&\textbf{38.6}&\textbf{38.6}&\textbf{38.6}&24s& 45.02\\
\rowcolor{green!40}\textbf{FCOS-421C(2x)} &30.4&77.8&16.6&30.4&41.9&41.9&41.9&41.9&23s &43.15\\
\rowcolor{green!40}\textbf{FCOS-421C(10x)}& \textbf{35.5}&\textbf{81.4}&\textbf{24.7}&\textbf{35.5}&\textbf{47.3}&\textbf{47.3}&\textbf{47.3}&\textbf{47.3}&\textbf{23s}&\textbf{43.15}\\
RetinaNet\cite{lin2017focal}& 18.1&57.0&5.5&18.1&36.9&36.9&36.9&36.9&24s&45.02 \\
\rowcolor{green!40}\textbf{RetinaNet-421C}&\textbf{18.6}&\textbf{57.0}&\textbf{6.2}&\textbf{18.6}&\textbf{37.1}&\textbf{37.1}&\textbf{37.1}&\textbf{37.1}&\textbf{23s}&\textbf{43.15}\\

Swin Transformer \cite{9710580}& 13.8&35.3&8.2&13.9&18.6&18.6&18.6&18.6&41s &76.92\\
\rowcolor{green!40}\textbf{Swin Transformer-421C}& \textbf{14.0}&\textbf{35.4}&\textbf{8.5}&\textbf{14.0}&\textbf{19.0}&\textbf{19.0}&\textbf{19.0}&\textbf{19.0}&43s& 80.68\\
\rowcolor{green!40}\textbf{Swin Transformer-421C(10x)} &\textbf{17.5}&\textbf{38.2}&\textbf{13.5}&\textbf{17.5}&\textbf{21.1}&\textbf{21.1}&\textbf{21.1}&\textbf{21.1}&\textbf{41s} &\textbf{76.92}\\

YOLOv7\cite{wang2022yolov7}&22.1&68.4&6.2&22.1&2.0&15.6&36.6&36.6&9s&16.88\\
\rowcolor{green!40}\textbf{YOLOv7-421C}&\textbf{26.6}&\textbf{75.5}&\textbf{10.1}&\textbf{26.6}&\textbf{2.3}&\textbf{18.1}&\textbf{40.2}&\textbf{40.2}&\textbf{9s}&\textbf{16.88}\\

RTMDet\cite{lyu2022rtmdet}&34.6&81.7&22.1&34.6&3.0&22.0&46.6&46.6&11s&20.63\\
\rowcolor{green!40}\textbf{RTMDet-421C}&\textbf{35.5}&\textbf{83.9}&\textbf{22.4}&\textbf{35.5}&\textbf{3.0}&\textbf{22.3}&\textbf{47.7}&\textbf{47.7}&\textbf{11s}&\textbf{20.63}\\

YOLOv8\cite{yolov8} &55.1&93.9&57.1&55.1&\textbf{3.7}&30.5&63.7&63.7&10s &18.76\\
\rowcolor{green!40}\textbf{YOLOv8-421C}&\textbf{55.5}&\textbf{94.2}&\textbf{57.6}&\textbf{55.5}&3.6&\textbf{30.9}&\textbf{64.4}&\textbf{64.4}&\textbf{9s}&\textbf{16.89}\\
\hline
\end{tabular}
\end{table*}

\begin{table}
\renewcommand{\arraystretch}{1.5}
\setlength\tabcolsep{3pt}
\caption{AP and AR of YOLOv7 with economical and complete Renormalized Connection methods: C421C and E421C on IPIU test set.}
\label{table_APAR_ipiu_c421c_vs_e421c}
\centering
\begin{tabular}{llllll}
\hline
\rowcolor{gray!40}Method & $AP^{0.5:0.95}$ & $AP^{0.5}$ & $AP^{0.75}$ & $AP^{S}$ &T(s) ${\downarrow}$\\
\hline
C421C&24.0&71.8&7.3&24.0&12s\\
\rowcolor{green!40}E421C&\textbf{26.6}&\textbf{75.5}&\textbf{10.1}&\textbf{26.6}&\textbf{9s}\\
\hline
\rowcolor{gray!40}Method& $AR_{1}$ & $AR_{10}$ & $AR_{100}$ & $AR_{100}^{S}$ &T/f (ms) ${\downarrow}$ \\
\hline
C421C&2.0&16.9&38.3&38.3&22.51\\
\rowcolor{green!40}E421C&\textbf{2.3}&\textbf{18.1}&\textbf{40.2}&\textbf{40.2}&\textbf{16.88}\\
\hline
\end{tabular}
\end{table}

\subsection{Results and Discussion}
\subsubsection{Results on IPIU}
The performance comparisons between the detectors embedded with the proposed economic Renormalized Connection and the state-of-the-art approaches on the IPIU test set are shown in Table \ref{table_APAR_ipiu}. For simplicity, we use 421C to represent the economic renormalized Connection that satisfies the scaling invariance property. In the complete connection experiments, we denote the economical 421 Connection and the complete 421 Connection by E421C and C421C, respectively. We designed KDN to generate and implement RC on single-branch detectors, and the amplification factors $(\lambda_2, \lambda_3, \lambda_4)$ of the KDN satisfy the ratio of 4:2:1. Therefore, we can use 421C to denote the RC of KDN on single-branch detectors. From the results, we can draw the following observations.
\begin{itemize}
\item[i)]
As shown in Table \ref{table_APAR_ipiu} and Fig. \ref{heatmap_comparison} (a), the simplest form of RC performs well in the difficult scale-preferred task of tiny moving vehicle detection and drastically reduces inference activations of irrelevant feature branches in FPN. The economical RC acting with a focal loss would further eliminate negative samples and focus more on the hard negative samples generated by $P_3$. In other words, the RC implements the ``divide and conquer'' idea of FPN not only for information flow renormalization and learning redirection for the hard scale-preferred task.
\end{itemize}

\begin{itemize}
\item[ii)]
In the first row of Table \ref{table_APAR_ipiu}, the mAP of Faster R-CNN-421C (KDN) reaches 26.3\%, which exceeds a large fraction of powerful multi-branch detectors. This result verifies that KDN is an efficient and effective feature renormalization method, realizing the application of renormalization group theory in feature extraction.
\end{itemize}

\begin{itemize}
\item[iii)]
By carefully observing the experimental results, anchor-based and anchor-free methods perform differently in detecting tiny objects. The best performer among the anchor-based methods is YOLOv7 with an $AP^{0.5:0.95}$ of 22.1\%. RetinaNet achieves $AP^{0.5:0.95}$ of 18.1\% after schedule\_1x training, while FCOS achieves $AP^{0.5:0.95}$ of 26.0\%. With further training, the AP of FCOS improves significantly, reaching 35.5\% after 120 epochs. The performance of Swin Transformer is at the same level as Cascade R-CNN. Compared with Swin Transformer-421C, the AP of HTC-421C is improved by 1.8\%. The powerful YOLOv8 achieves the highest AP of 55.1\%, benefiting from the powerful PAFPN and anchor-free algorithms. 421C improves the $AP^{0.5:0.95}$ of YOLOv8 by another 0.4\%. 
\end{itemize}

\begin{itemize}
\item[iv)]
The inference time for the IPIU test set also varies with the network structure. Cascaded refinement increases the inference time, as in the case of Cascade R-CNN and HTC baselines. The inference times for FCOS and RetinaNet are similar, both being $43-45\,ms/frame$. Swin Transformer's mask branch slows down the prediction. YOLOv8-421C is the fastest, at $16.89\,ms/frame$.
\end{itemize}

\begin{figure}
\centering
\includegraphics[width=3.5in]{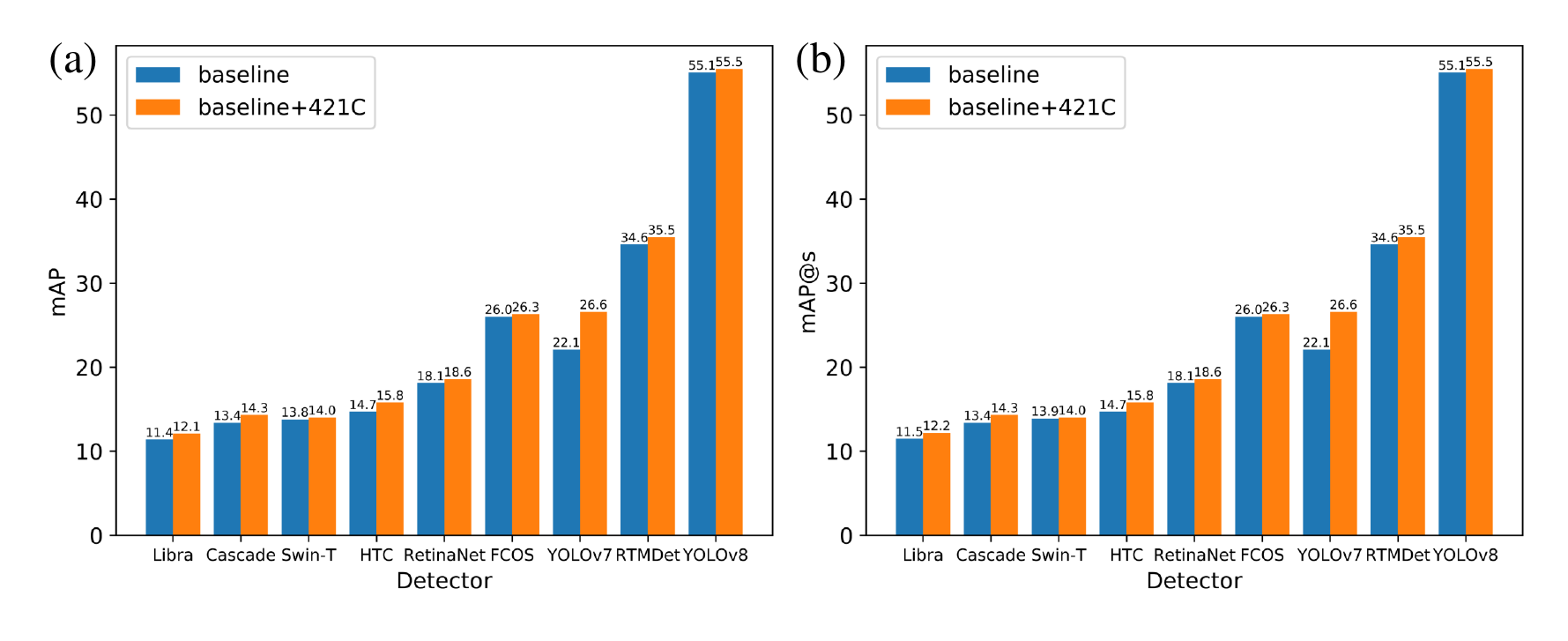}
\caption{$AP^{0.5:0.95}$ (\%) (a) and $AP^{S}$ (\%) (b) comparison between 421C embedded detectors (orange) and baselines (blue) on IPIU test set.}
\label{ipiu_AP_plots}
\end{figure}

\begin{itemize}
\item[v)]
The highest $AP^{0.5:0.95}$ and $AP^{S}$ for baselines and 421C on 9 detectors are plotted in Fig. \ref{ipiu_AP_plots}. We can see that 421C consistently achieves higher accuracy on all baselines. This proves that 421C does improve the accuracy of different types of high-performance detectors.
\end{itemize}

\begin{itemize}
\item[vi)]
As shown in Table \ref{table_APAR_ipiu_c421c_vs_e421c}, we can see that the complete form degrades the performance in terms of both mAP and inference speed with the addition of RCs to all branches. This further demonstrates that the economical form of RC can achieve feature renormalization for FPN-based multi-branch detectors in difficult scale-preferred tasks.
\end{itemize}

\begin{table*}
\renewcommand{\arraystretch}{1.5}
\setlength\tabcolsep{3pt}
\caption{AP and AR of single-branch detectors Faster R-CNN and R-fcn with Renormalized Connection (RC) that is the implementation of KDN, all baselines, and SSD-FPN on RSOD Aircraft Val set. AP: $AP^{0.5:0.95}$.}
\label{table_AP_rsod_aircraft_1}
\centering
\begin{tabular}{lllllllllllllll}
\hline
\rowcolor{gray!40}Method& Backbone &$AP$ & $AP^{0.5}$ & $AP^{0.75}$ & $AP^{S}$ & $AP^{M}$ & $AP^{L}$&$AR_{1}$ & $AR_{10}$ & $AR_{100}$ & $AR_{100}^{S}$ & $AR_{100}^{M}$&$AR_{100}^{L}$ &T(ms) ${\downarrow}$\\
\hline
Faster R-CNN\cite{faster_rcnn}&R50&47.3&87.6&45.2&14.5&51.4&\textbf{74.1}
&5.9&40.8&57.4&25.2&61.5&\textbf{78.4}
&29.9\\
\rowcolor{green!40}\textbf{Faster R-CNN-421C}& R50&\textbf{56.6}&\textbf{89.5}&\textbf{66.4}&\textbf{24.3}&\textbf{61.1}&72.2
&\textbf{6.1}&\textbf{43.9}&\textbf{62.6}&\textbf{32.3}&\textbf{67.3}&75.2
&\textbf{28.7}\\

Faster R-CNN&R101&52.0&88.7&55.6&19.8&56.6&\textbf{75.3}
&5.8&\textbf{43.5}&61.6&30.8&65.8&\textbf{79.9}
&29.1\\
\rowcolor{green!40}\textbf{Faster R-CNN-421C}&R101&\textbf{56.0}&\textbf{91.2}&\textbf{65.5}&\textbf{26.5}&\textbf{60.1}&69.2
&\textbf{5.9}&43.4&\textbf{62.4}&\textbf{35.0}&\textbf{66.7}&72.8
&\textbf{27.6}\\

R-fcn\cite{dai2016r}&R50&41.4&81.2&35.9&11.6&44.5&64.4
&5.4&36.5&49.4&18.7&53.1&\textbf{70.8}
&28.7\\
\rowcolor{green!40}\textbf{R-fcn-421C}&R50&\textbf{52.4}&\textbf{86.4}&\textbf{59.0}&\textbf{20.8}&\textbf{57.2}&\textbf{63.7}
&\textbf{5.9}&\textbf{41.9}&\textbf{59.0}&\textbf{29.2}&\textbf{64.2}&66.3
&\textbf{27.1}\\

R-fcn&R101&41.8&80.5&40.0&13.8&44.8&64.3
&5.3&35.7&49.9&20.3&53.5&\textbf{70.8}
&29.7\\
\rowcolor{green!40}\textbf{R-fcn-421C}&R101&\textbf{52.8}&\textbf{86.8}&\textbf{60.6}&\textbf{21.2}&\textbf{57.5}&\textbf{65.8}
&\textbf{5.9}&\textbf{42.3}&\textbf{60.0}&\textbf{33.3}&\textbf{64.3}&69.5
&\textbf{29.3}\\

SSD-FPN\cite{lin2017focal}&R50&41.9&71.5&44.3&12.0&46.4&55.0
&5.8&36.4&52.9&24.1&57.8&60.9
&26.2\\
\hline
\end{tabular}
\end{table*}

\begin{table*}
\renewcommand{\arraystretch}{1.5}
\setlength\tabcolsep{3pt}
\caption{AP and AR of detectors with 421C and baselines on RSOD Aircraft Val set. Image size: ${1333}\times{800}$, except for YOLOv8 ${1280}\times{1280}$. Backbone: ResNet-50 for ConvNet detectors, Swin Transformer: Swin-T, YOLOv8: CSPDarknet, Centernet: ResNet-18, and CornerNet: HourglassNet-104. 421 Connection added on the top-down (421C (A)) and bottom-up path (421C (B)) of PAFPN.}
\label{table_AP_rsod_aircraft_2}
\centering
\begin{tabular}{lllllllllllllll}
\hline
\rowcolor{gray!40}Method & $AP$ & $AP^{0.5}$ & $AP^{0.75}$ & $AP^{S}$ & $AP^{M}$ & $AP^{L}$
& $AR_{1}$ & $AR_{10}$ & $AR_{100}$ & $AR_{100}^{S}$ & $AR_{100}^{M}$&$AR_{100}^{L}$
&T (s) ${\downarrow}$&T/f (ms) ${\downarrow}$\\
\hline
Cascade R-CNN\cite{cai2019cascade}&65.9&95.2&80.7&41.5&68.3&\textbf{78.4}
&70.5&70.5&70.5&50.4&73.3&81.1
&10s&131.57\\
\rowcolor{green!40}\textbf{Cascade R-CNN-421C}&\textbf{66.8}&95.4&\textbf{84.4}&44.0&\textbf{69.5}&78.2
&71.6&71.6&71.6&50.3&74.6&82.9
&\textbf{10s}&131.57\\

Cascade+GRoIE \cite{rossi2021novel} &66.0&\textbf{96.1}&81.8&44.1&68.3&75.3
&71.1&71.1&71.1&53.1&73.8&78.9
&22s&289.47\\
\rowcolor{green!40}\textbf{Cascade+GRoIE-421C}& 65.5&95.3&81.2&\textbf{44.2}&67.6&77.2
&70.1&70.1&70.1&51.8&72.5&80.9
&23s&302.63\\

Faster R-CNN\cite{fpn}& 63.0&95.2&76.5&35.8&65.7&\textbf{76.9}
& 68.3&68.3&68.3&47.5&71.0&81.4
&6.6s&86.84\\
\rowcolor{green!40}\textbf{Faster R-CNN-421C}& \textbf{64.2}&\textbf{95.4}&\textbf{79.6}&\textbf{43.6}&\textbf{66.6}&74.1
&69.6&69.6&69.6&51.7&72.3&77.7
&6.9s&90.78\\

Grid R-CNN\cite{lu2019grid}&66.1&95.2&81.1&43.0&68.5&\textbf{78.2}
& 70.6&70.6&70.6&50.4&73.4&82.0
&8.4s&110.53\\
\rowcolor{green!40}\textbf{Grid R-CNN-421C}&\textbf{66.4}&\textbf{95.9}&\textbf{82.4}&\textbf{44.5}&\textbf{68.4}&77.7
& 71.4&71.4&71.4&53.8&73.6&82.4
&9s&118.42\\

HTC\cite{chen2019hybrid}& 66.7&95.8&\textbf{83.5}&42.9&69.2&79.0
&71.5&71.5&71.5&51.9&74.1&83.3
&15s&197.36\\
\rowcolor{green!40}\textbf{HTC-421C}& \textbf{67.6}&\textbf{95.8}&82.9&\textbf{43.0}&\textbf{69.8}&\textbf{80.5}
&72.1&72.1&72.1&53.3&74.5&84.4
&\textbf{15s}&197.36\\

Sparse R-CNN\cite{9577670}& 34.8&61.4&36.4&\textbf{15.5}&37.9&52.8
& 58.7&58.7&58.7&26.6&63.2&75.4
&6s&78.94\\
\rowcolor{green!40}\textbf{Sparse R-CNN-421C}& \textbf{36.8}&\textbf{62.5}&\textbf{42.2}&13.4&\textbf{40.8}&\textbf{54.6}
& 57.4&57.4&57.4&22.6&62.0&79.4
&7s&92.10\\

Mask R-CNN\cite{he2017mask}& 64.7&95.1&\textbf{81.1}&41.8&67.2&\textbf{77.3}
& 69.7&69.7&69.7&50.0&72.4&81.0
&9s&78.94\\
\rowcolor{green!40}\textbf{Mask R-CNN-421C}& \textbf{65.3}&\textbf{95.1}&80.6&\textbf{42.9}&\textbf{67.7}&77.2& 70.1&70.1&70.1&51.2&72.6&81.3
&10s&131.57\\

PANet\cite{liu2018path}& 64.0&95.1&78.9&41.0&66.4&74.9
& 68.7&68.7&68.7&51.0&71.2&78.4
&7s&92.10\\
\rowcolor{green!40}\textbf{PANet-421C(A)}&
64.7&\textbf{95.2}&\textbf{81.0}&\textbf{43.1}&67.1&\textbf{77.4}
&70.1&70.1&70.1&50.8&72.6&81.4
&\textbf{7s}&92.10\\
\rowcolor{green!40}\textbf{PANet-421C(B)}&
\textbf{65.0}&94.2&80.7&42.1&\textbf{67.5}&77.0
&69.7&69.7&69.7&49.7&72.5&81.1
&\textbf{7s}&92.10\\

Libra R-CNN\cite{pang2019libra}& 64.1&\textbf{95.6}&77.7&39.2&66.8&77.2
& 69.3&69.3&69.3&51.0&71.6&81.1
&7s&92.10\\
\rowcolor{green!40}\textbf{Libra R-CNN-421C}& \textbf{65.4}&95.0&\textbf{81.7}&\textbf{43.6}&\textbf{67.5}&\textbf{77.7}
& 70.5&70.5&70.5&53.7&72.7&81.4
&\textbf{7s}&92.10\\
\hline
\hline

FCOS\cite{9229517}& 51.7&91.2&52.6&23.1&56.1&61.3
& 58.8&58.8&58.8&35.1&62.7&66.1
&6s&78.94\\
\rowcolor{green!40}\textbf{FCOS-421C}& \textbf{54.3}&\textbf{91.6}&\textbf{60.3}&\textbf{23.2}&\textbf{59.2}&\textbf{65.5}
& 60.9&60.9&60.9&33.7&65.3&70.4
&\textbf{6s}&78.94\\

Centernet\cite{duan2019centernet}& 46.7&93.1&38.9&17.7&48.4&66.7
& 51.8&51.8&51.8&28.4&54.3&70.6
&3.1s&40.78\\
CornerNet\cite{cornetnet}& 48.4&68.0&58.5&17.6&64.7&56.1
& 68.8&68.8&68.8&46.5&71.6&83.9
&72.4s&952.63\\
YOLOF \cite{chen2021you}& 45.5&82.9&45.5&9.9&50.5&63.5
& 54.7&54.7&54.7&17.7&60.3&69.9
&4s&52.63\\
\hline
\hline
DETR\cite{carion2020end}&56.1&90.7&65.3&19.2&60.8&68.8
&64.4&64.4&64.4&32.7&69.4&76.2
&6.4s&84.21\\

Swin Transformer\cite{9710580}& 63.6&\textbf{96.1}&78.2&42.6&65.8&74.7
& 68.2&68.2&68.2&50.1&70.7&77.8
&12s&157.89\\
\rowcolor{green!40}\textbf{Swin Transformer-421C}& \textbf{64.1}&95.9&\textbf{78.5}&\textbf{44.6}&\textbf{66.0}&\textbf{75.9}
& 69.2&69.2&69.2&53.0&71.2&80.0
&12.4s&163.15\\
\hline
\hline
RTMDet\cite{lyu2022rtmdet}&71.5&97.4&\textbf{89.8}&51.4&73.1&\textbf{83.5}
&6.6&50.7&75.6&60.3&77.5&86.8&2s&26.31\\
\rowcolor{green!40}\textbf{RTMDet-421C}&\textbf{71.5}&\textbf{97.5}&89.4&\textbf{52.4}&\textbf{73.1}&82.7
&6.6&50.8&75.7&60.3&77.6&86.1&\textbf{2s}&26.31\\
YOLOv7\cite{wang2022yolov7}&66.4&97.1&81.2&44.5&\textbf{68.4}&79.0
&6.6&\textbf{48.6}&\textbf{72.3}&\textbf{58.4}&73.8&\textbf{83.8}&1s&13.15\\
\rowcolor{green!40}\textbf{YOLOv7-421C}&\textbf{66.5}&\textbf{97.2}&\textbf{82.0}&\textbf{45.4}&68.2&\textbf{79.0}
&\textbf{6.6}&48.4&72.2&57.6&\textbf{73.9}&82.3&\textbf{1s}&13.15\\

YOLOv8\cite{yolov8}&71.0&\textbf{97.1}&\textbf{88.1}&\textbf{52.3}&72.2&82.9
&6.8&50.8&76.0&62.1&77.6&86.7
&1s&13.15\\
\rowcolor{green!40}\textbf{YOLOv8-421C(A)}&\textbf{71.5}&\textbf{97.1}&87.4&51.7&72.6&\textbf{84.6}
&6.8&51.1&75.8&61.1&77.4&87.3
&\textbf{1s}&13.15\\
\rowcolor{green!40}\textbf{YOLOv8-421C(B)}&\textbf{71.5}&\textbf{97.1}&86.3&52.1&\textbf{73.1}&82.6
&6.7&51.0&75.9&61.6&77.5&86.3
&\textbf{1s}&13.15\\
\hline
\end{tabular}
\end{table*}

\begin{table}
\renewcommand{\arraystretch}{1.5}
\setlength\tabcolsep{3pt}
\caption{AP and AR of Swin Transformer Mask R-CNN with two types of complete connection methods: C421C and weighted Complete Connection (WeightedCC) on RSOD Aircraft Val set. Backbone: Swin-T.}
\label{table_APAR_rsod_aircraft_coupling_modes}
\centering
\begin{tabular}{lllllll}
\hline
\rowcolor{gray!40}Extractor & $AP^{0.5:0.95}$ & $AP^{0.5}$ & $AP^{0.75}$ & $AP^{S}$ & $AP^{M}$ & $AP^{L}$\\
\hline
\rowcolor{green!40}C421C&\textbf{64.5}&96.0&\textbf{80.1}&42.7&\textbf{66.6}&\textbf{76.7}\\

WeightedCC&64.1&\textbf{96.8}&78.1&\textbf{43.2}&66.2&76.6\\
\hline
\rowcolor{gray!40}Extractor& $AR_{1}$ & $AR_{10}$ & $AR_{100}$ & $AR_{100}^{S}$ & $AR_{100}^{M}$&$AR_{100}^{L}$ \\
\hline
\rowcolor{green!40}C421C&69.0&69.0&69.0&50.8&71.4&80.1\\
WeightedCC&69.1&69.1&69.1&52.0&71.3&80.8\\
\hline
\end{tabular}
\end{table}

\subsubsection{Results on RSOD Aircraft}
The comparison of AP and AR of the Renormalized Connection and baselines on single-branch detectors (KDN) and multi-branch detectors on RSOD Aircraft are summarized in Table \ref{table_AP_rsod_aircraft_1} and Table \ref{table_AP_rsod_aircraft_2}, respectively. The results of the Complete 421 Connection and a weighted complete connection method are listed in Table \ref{table_APAR_rsod_aircraft_coupling_modes}. 
\begin{itemize}
\item[i)]
As shown in Table~\ref{table_AP_rsod_aircraft_1}, KDN far outperforms the baseline on almost all criteria in two typical classical single-branch detectors (Faster R-CNN and R-fcn). The Renormalized Connection improves the $AP^{0.5:0.95}$ of Faster R-CNN by 9.3\% on ResNet-50 and 4\% on ResNet-101. Among all the R-fcn experiments, 421C has the most significant improvement in $AP^{0.5:0.95}$ with 11\%. The RC is particularly effective for small object detection on the shallower backbone ResNet-50, exceeding the $AP^{S}$ of Faster R-CNN by 9.8\% (from 14.5\% to 24.3\%) and R-fcn by 9.2\% (from 11.6\% to 20.8\%). There has been a consistent improvement in the accuracy of small and medium objects across all experiments. The results confirm that KDN achieves renormalization of features for single-branch detectors on small object detection tasks.
\end{itemize}
\begin{figure}
\centering
\includegraphics[width=3.5in]{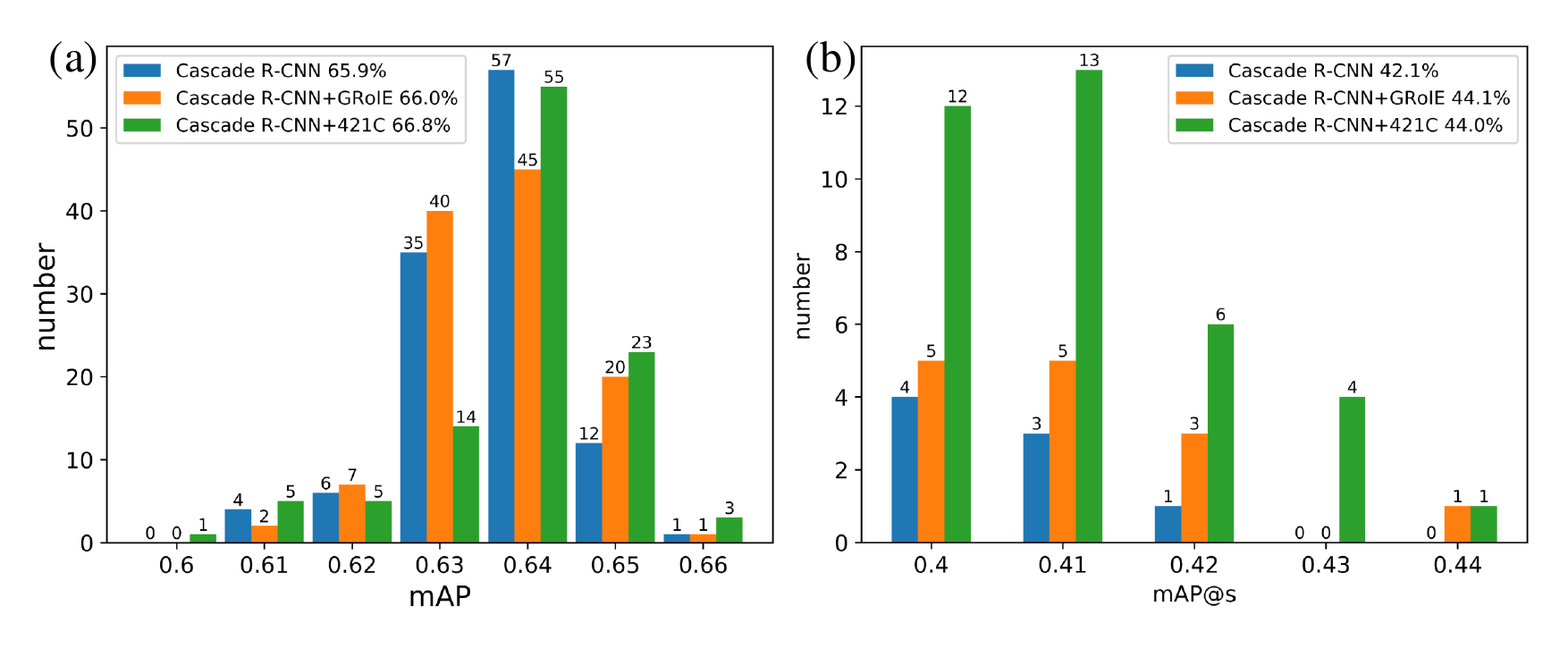}
\caption{RSOD Aircraft Val set inference results of $AP^{0.5:0.95}$ (a) and $AP^{S}$ (b) Statistical bar plots of Cascade R-CNN (blue), Cascade R-CNN+GRoIE (orange), and Cascade R-CNN-421C (green). After each training epoch, the 3 detectors are evaluated on the whole Val set to obtain results. Each interval contains an incremental AP of 0.01. For example, the first green bar in (a) means that there is one $AP^{0.5:0.95}$ result that falls in [0.60, 0.61].}
\label{rsod_aircraft_AP_plots}
\end{figure}
\begin{itemize}
\item[ii)]
As shown in Fig. \ref{heatmap_comparison} (b), the simplest form of renormalized connection, E421C, performs well in the scale-preferred task of small aircraft detection, drastically reducing the inference activations of irrelevant feature branch ($P_5$) in FPN. The economical renormalized connection with focal loss function will further eliminate negative samples and focus more on the hard negative samples generated by $P_3$. That is, in the RSOD Aircraft scale-preferred task, the Renormalized Connection implements the ``divide-and-conquer'' idea of FPN and observably rearranges the subtask assignment.
\end{itemize}

\begin{itemize}
\item[iii)]
Table \ref{table_AP_rsod_aircraft_2} lists the AP and AR results of 13 well-designed detectors with multi-branch detectors inserted with 421C and 17 baselines, including 6 powerful R-CNN-based two-stage baselines, 3 instance segmentation baselines, 3 anchor-free one-stage baselines, 2 transformer baselines, 1 combined with RoI extractor, and 3 real-time detectors (including the state-of-the-art YOLOv8). With all robust baselines, all 421C experiments achieve performance gains on most of the evaluated metrics. The 421C does not add any additional trainable layers (convolution/multi-head attention) but performs well on different types of detectors, demonstrating the importance of enriching features. Compared to Cascaded R-CNN, 421C improves $AP^{0.5:0.95}$ by 0.9\% and $AP^{S}$ by 2.5\%. Combining 421C with the GRoIE RoI extractor increases $AP^{S}$ by 0.2\% but decreases $AP^{0.5:0.95}$ by 1.3\%. YOLOv8-421C achieves the highest AP of 71.5\%, exceeding the powerful baseline by 0.5\%.
\end{itemize}

\begin{itemize}
\item[iv)]
In Section \ref{f421c}, we introduced the Complete 421 Connection method and the Weighted Complete Connection method. As shown in Table \ref{table_APAR_rsod_aircraft_coupling_modes}, C421C on Swin Transformer Mask R-CNN ($AP^{0.5:0.95}$: 64.5\%, $AP^{S}$: 42.7\%, $AP^{M}$: 66.6\%, $AP^{L}$: 76.7\%) further improves the accuracy of E421C ($AP^{0.5:0.95}$: 64.1\%, $AP^{S}$: 44.6\%, $AP^{M}$: 66.0\%, $AP^{L}$: 75.9\%). While C421C has the same AP as E421C (64.1\%), it drops $AP^{S}$ by 1.4\%. Weighted Complete Connection gets an $AP^{S}$ of 43.2\%, which is 0.5\% higher than C421C, but 1.4\% lower than E421C. And it has the same AP as E421C. All observations show that the simplest economical Renormalized Connection is the economical choice for small object detection.
\end{itemize}

\begin{table*}
\renewcommand{\arraystretch}{1.5}
\setlength\tabcolsep{3pt}
\caption{AP and AR of detectors with $n21$C and baselines on RSOD Val set. Image size: ${1333}\times{800}$. YOLOv8: ${1280}\times{1280}$. Backbone: ResNet-50 for ConvNet detectors, Swin-T for Swin Transformer, and YOLOv8CSPDarknet for YOLOv8. AP: $AP^{0.5:0.95}$.}
\label{table_AP_rsod}
\centering
\begin{tabular}{lllllllllllllll}
\hline
\rowcolor{gray!40}Method & $AP$ & $AP^{0.5}$ & $AP^{0.75}$ & $AP^{S}$ & $AP^{M}$ & $AP^{L}$& $AR_{1}$ & $AR_{10}$ & $AR_{100}$ & $AR_{100}^{S}$ & $AR_{100}^{M}$&$AR_{100}^{L}$ &T(s) ${\downarrow}$&T/f (ms) ${\downarrow}$\\
\hline
Cascade R-CNN\cite{cai2019cascade}&62.9&93.3&71.3&40.6&70.3&66.6&68.6&68.6&68.6&50.9&76.0&71.8&18s&113.92\\
\rowcolor{green!40}\textbf{Cascade R-CNN-421C}&64.6&92.8&75.1&42.1&70.3&68.8&70.2&70.2&70.2&49.0&75.7&73.8&19s&120.25\\

HTC\cite{chen2019hybrid}& 64.2&92.6&73.6&40.9&71.3&67.6&70.4&70.4&70.4&50.1&76.6&73.5&37s&234.18\\
\rowcolor{green!40}\textbf{HTC-421C}& 64.2&92.0&72.9&43.1&71.6&67.8&71.4&71.4&71.4&52.6&77.3&74.7&35s&221.52\\

Swin Transformer\cite{9710580}& 65.3&95.8&73.4&41.8&70.1&68.7&70.2&70.2&70.2&49.7&75.1&73.3&23s&145.57\\
\rowcolor{green!40}\textbf{Swin Transformer-421C}& 66.1&95.8&72.3&41.1&70.1&70.0&70.7&70.7&70.7&49.0&75.3&74.1&22s&139.24\\

RTMDet\cite{lyu2022rtmdet}&59.4&86.7&67.5&49.0&71.0&63.1&28.3&58.4&69.5&58.3&78.2&72.6&7s&44.3\\
\rowcolor{green!40}\textbf{RTMDet-421C}&59.7&85.9&69.3&47.1&71.4&63.4&27.9&59.0&70.8&56.3&78.8&73.8&6s&37.9\\

YOLOv7\cite{wang2022yolov7}&62.7&90.4&70.9&50.7&71.8&66.3
&31.2&60.5&70.8&61.1&79.0&73.7&4s&25.31\\
\rowcolor{green!40}\textbf{YOLOv7-421C}&62.9&91.4&72.1&48.6&71.8&66.4
&32.3&60.7&71.4&59.5&78.8&74.0&3s&18.98\\
\rowcolor{green!40}\textbf{YOLOv7-521C}
&63.0&90.7&71.3&49.6&72.6&66.5
&32.7&60.7&71.4&59.9&78.2&74.3&3s&18.98\\
YOLOv8\cite{yolov8}&\emph{68.4}&93.1&77.9&50.1&72.5&71.7
&36.0&65.1&75.7&60.1&78.5&78.6&4s&25.31\\
\rowcolor{green!40}\textbf{YOLOv8-421C}&\textbf{68.7}&93.5&79.6&49.6&74.3&71.8&36.2&64.2&74.4&60.3&79.7&77.2&4s&25.31\\
\rowcolor{green!40}\textbf{YOLOv8-521C}&68.7&93.6&76.5&50.4&73.9&72.4
&35.6&64.8&75.2&60.3&80.0&78.4&4s&25.31\\
YOLOv8 Top$AP^{S}$\cite{yolov8}&67.0&92.1&76.7&\emph{51.9}&73.8&70.0&34.6&64.4&75.4&\emph{61.7}&79.4&78.0&4s&25.31\\
\rowcolor{green!40}\textbf{YOLOv8-421C} Top$AP^{S}$&67.3&91.9&75.0&\textbf{52.6}&74.9&70.7&34.6&64.3&75.4&\textbf{62.4}&79.7&78.3&4s&25.31\\
\rowcolor{green!40}\textbf{YOLOv8-521C} Top$AP^{S}$&
68.7&92.5&77.4&51.8&74.9&72.3
&35.6&65.6&76.4&61.0&79.8&79.7&4s&25.31\\
\hline
\end{tabular}
\end{table*}

\begin{itemize}
\item[v)]
From the ``T'' and ``T/f'' columns in Table \ref{table_AP_rsod_aircraft_2}, the fastest algorithm is YOLOv8. YOLOv8 and RTMDet enable real-time detection. Mask prediction slows down the inference process. Portable networks predict fast. Overall, the inference time of the detectors is acceptable.
\end{itemize}

\begin{itemize}
\item[vi)]
Fig. \ref{rsod_aircraft_AP_plots} shows the AP results of the baseline, GRoIE, and 421C. During training, 421C has higher AP scores than GRoIE, with 81 scores higher than 63.0\% out of a total of 120 results. Especially in $AP^{S}$, 421C has 36 scores higher than 40.0\%, which is much higher than the other two methods.
\end{itemize}
\subsubsection{Results on RSOD}
The comparison of the performance of a series of $n21$C and the corresponding baseline on RSOD is reported in Table \ref{table_AP_rsod}. The best results are shown in bold. The $n21$C has been inserted into Cascade R-CNN, HTC, Swin Transformer, RTMDet, YOLOv7, and YOLOv8. Each baseline represents a typical structure in the CV task. From these results, we have the following observations.
\begin{itemize}
\item[i)]
As shown in Table \ref{table_AP_rsod}, $n21$Cs outperform all strong baselines in terms of $AP^{0.5:0.95}$. The AP scores of $n21$C are competitive with a variety of well-designed networks, Cascade R-CNN, Swin Transformer, and YOLOv8. Specifically, the $AP^{0.5:0.95}$ of 421C improves by 1.7\%, 0.8\%, and 0.3\%, respectively, when compared to the best AP score of the three baselines mentioned above, proving the effectiveness of $n21$C across a variety of modern detectors. These validate the good renormalization effect of the economical $n21$C on the task with diverse scale distributions, a multi-class satellite object detection dataset.
\end{itemize}

\begin{figure}
\centering
\includegraphics[width=3.5in]{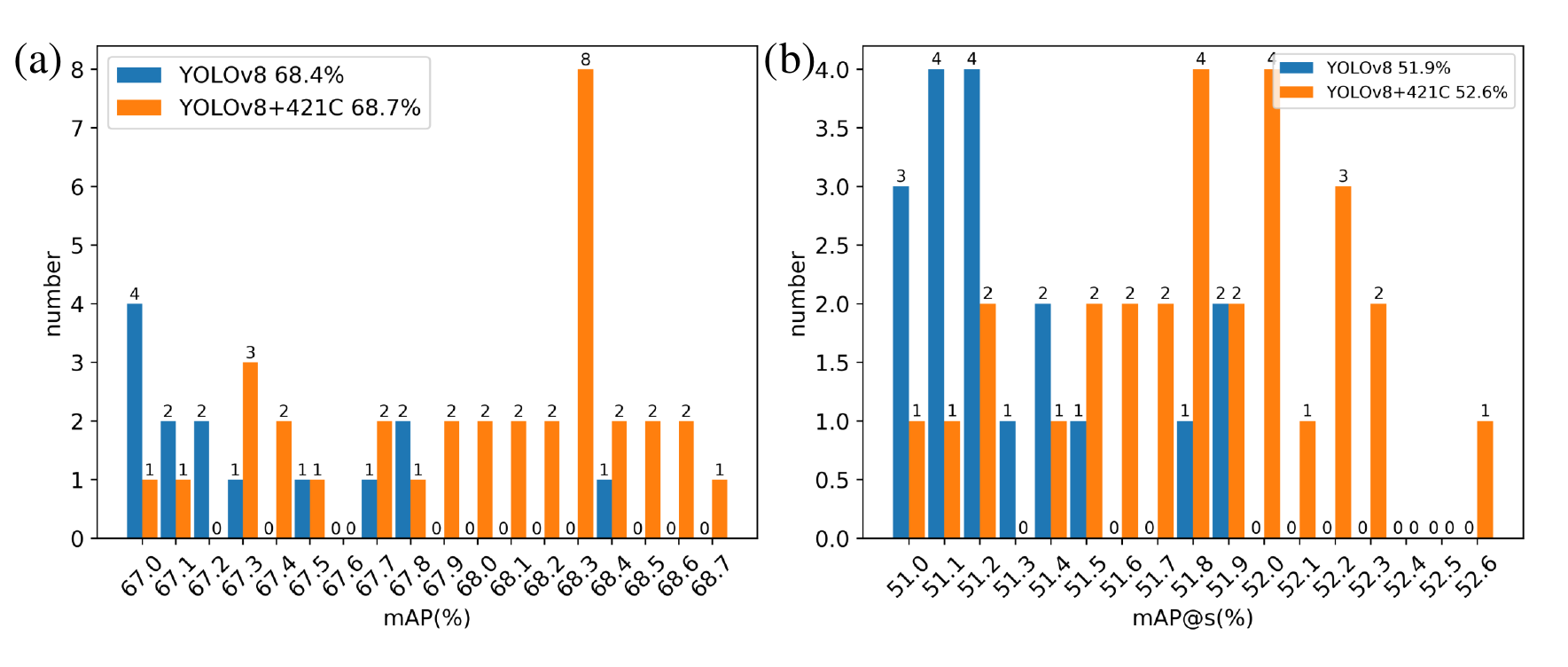}
\caption{RSOD validation set inference results of $AP^{0.5:0.95}$ (a) and $AP^{S}$ (b) Statistical bar plots of YOLOv8 (blue) and YOLOv8-421C (orange). After each training epoch, the two networks are evaluated on the entire Val set to obtain the results. Each label in the $x$ axis represents an AP value in \%.}
\label{rsod_AP_plots}
\end{figure}

\begin{itemize}
\item[ii)]
Let us see how the 421C performs on small objects. As shown in the ``$AP^{S}$'' column, the 421C outperforms the powerful HTC by 2.2\% in $AP^{S}$, 0.3\% in $AP^{M}$, and 0.2\% in $AP^{L}$. In $AP^{S}$, the 421C outperforms the Cascade R-CNN by 1.5\%. The last three rows show the highest $AP^{S}$ for YOLOv8-based PAFPN, 421C and 521C, with 421C achieving 0.7\% improvement in $AP^{S}$, 1.1\% in $AP^{M}$, 0.7\% in $AP^{L}$, and 0.3\% in AP; and 521C achieving the highest $AP^{S}$ score of 52.6\%.
\end{itemize}

\begin{itemize}
\item[iii)]
The inference times are listed in the last two columns. YOLOv8-$n21$C achieves a real-time detection of 25.31ms per frame. The 421C reduces the total inference time of HTC and Swin Transformer by 2s and 1s respectively. This suggests that $n21$C is a lightweight and fast extractor, with no time overhead on the baselines.

\end{itemize}
\begin{itemize}
\item[iv)]
The $AP^{0.5:0.95}$ and $AP^{S}$ statistics of YOLOv8 and YOLOv8-421C are plotted in Fig. \ref{rsod_AP_plots}. From the plots, it can be seen that 421C consistently maintains high accuracy during the training process and achieves the highest AP. This proves that the 421C can continuously improve its performance on tasks with diverse scale distributions.
\end{itemize}

\begin{table*}
\renewcommand{\arraystretch}{1.5}
\setlength\tabcolsep{3pt}
\caption{AP of KDN experiments on MS COCO TOD val set. AP: $AP^{0.5:0.95}$.}
\label{table_AP_KDN_tod}
\centering
\begin{tabular}{llllllllllllll}
\hline
\rowcolor{gray!40}Method & Extractor & Image size & Steps & $AP$ & $AP^{0.5}$ & $AP^{0.75}$ & $AP^{S}$ & $AP^{M}$ & $AR_{1}$ & $AR_{10}$ & $AR_{100}$ & $AR_{100}^{S}$ & $AR_{100}^{M}$\\
\hline

R-fcn\cite{dai2016r}& ResNet101&${512}\times{300}$
&1.2M &7.5e-2&0.22&4.3e-2&7.1e-2&0.21
&7.5e-2&0.22&4.3e-2&7.1e-2&0.21\\
\hline

SSD-FPN\cite{lin2017focal}&ResNet50-FPN&${640}\times{640}$&1.2M &2.9e-2&1.1e-2&1.1e-2&3.1e-2&1.0e-2
&2.9e-2&1.1e-2&1.1e-2&3.1e-2&1.0e-2\\
SSD-FPN\cite{lin2017focal}&ResNet101-FPN&${640}\times{640}$&1.1M &1.5e-7&4.4e-7&2.2e-8&7.9e-7&5.0e-8
&1.3e-6&2.0e-5&3.1e-5&1.4e-5&1.9e-4\\
\hline
NASNet-A\cite{8579005}&ResNet101-NAS & ${300}\times{300}$&713&0&0&0&0&0&0&0&0&0&0\\
\hline
Faster R-CNN-FPN\cite{fpn} &ResNet50-FPN& ${640}\times{640}$&1M&1.2&2.2&1.1&1.1&1.9
&3.9&8.2&9.3&8.9&13.4\\
\hline
Faster R-CNN\cite{faster_rcnn}&ResNet101-$C_3$ & ${512}\times{300}$&1M &1.5&3.9&0.8&1.3&4.2
&3.9&7.9&9.1&8.2&15.7\\

\rowcolor{green!40}\textbf{Faster R-CNN-421C}&ResNet101-KDN & ${512}\times{300}$ & 1M &\textbf{3.9}&\textbf{7.6}&\textbf{3.3}&\textbf{3.8}&\textbf{6.5}
&\textbf{7.8}&\textbf{14.0}&\textbf{15.7}&\textbf{15.1}&\textbf{20.7}\\
\hline
\end{tabular}
\end{table*}

\begin{table*}
\renewcommand{\arraystretch}{1.5}
\setlength\tabcolsep{3pt}
\caption{AP and AR of detectors with 421C and baselines on MS COCO TOD-80 Val set. Image size: Cascade R-CNN: ${1333}\times{800}$; Swin Transformer: ${600}\times{600}$; YOLOv8: ${640}\times{640}$. Backbone: ResNet-50, Swin-T, and CSPDarknetResNet-50.} 
\label{table_AP_coco_tiny}
\centering
\begin{tabular}{llllllllllllll}
\rowcolor{gray!40}Method &Epoch& $AP^{0.5:0.95}$ & $AP^{0.5}$ & $AP^{0.75}$ & $AP^{S}$ & $AP^{M}$ & $AR_{1}$ & $AR_{10}$ & $AR_{100}$ & $AR_{100}^{S}$ & $AR_{100}^{M}$ &T(s) ${\downarrow}$&T/f (ms) ${\downarrow}$\\
\hline
Cascade R-CNN\cite{cai2019cascade}&33&8.2&13.7&8.8&8.4&\textbf{10.9}
&17.9&17.9&17.9&17.8&26.2&258s&119.22\\
\rowcolor{green!40}\textbf{Cascade R-CNN-421C}&33 &\textbf{8.9}&\textbf{14.6}&\textbf{9.8}&\textbf{9.3}&10.7
&\textbf{19.3}&\textbf{19.3}&\textbf{19.3}&\textbf{19.1}&30.8&262s&121.07\\
\hline
\rowcolor{green!40}\textbf{Swin Transformer-421C}&12&8.0&15.5&8.0&8.7&10.6
&18.7&18.7&18.7&18.8&31.2&180s&83.17\\
\hline
YOLOv8\cite{yolov8}&500&11.3&19.1&11.7&12.2&13.5
&15.7&29.3&33.7&33.7&46.8&46s&21.26\\
\rowcolor{green!40}\textbf{YOLOv8-421C}&500&\textbf{11.4}&18.4&\textbf{12.3}&\textbf{12.3}&\textbf{15.2}
&\textbf{16.0}&\textbf{29.6}&\textbf{34.2}&\textbf{33.7}&\textbf{48.0}&\textbf{46s}&\textbf{21.26}\\
\hline
\end{tabular}
\end{table*}

\begin{table*}
\renewcommand{\arraystretch}{1.5}
\setlength\tabcolsep{3pt}
\caption{AP and AR of detectors with 421C, 521C and baselines on NWPU VHR-10 test set. (Ablation study 1.4.) Image size: $1280\times1280$.}
\label{table_APAR_nwpu}
\centering
\begin{tabular}{lllllllllllllll}
\hline
\rowcolor{gray!40}Method & $AP$ & $AP^{0.5}$ & $AP^{0.75}$ & $AP^{S}$ & $AP^{M}$ & $AP^{L}$
& $AR_{1}$ & $AR_{10}$ & $AR_{100}$ & $AR_{100}^{S}$ & $AR_{100}^{M}$&$AR_{100}^{L}$
&T (s) ${\downarrow}$&T/f (ms) ${\downarrow}$\\
\hline
YOLOv7\cite{wang2022yolov7}&46.6&85.0&45.4&18.0&44.1&50.1
&20.6&50.1&57.9&31.3&57.0&55.9&3s&15.3\\
\rowcolor{green!40}\textbf{YOLOv7-421C}
&47.1&84.4&45.4&19.8&\textbf{45.6}&48.5
&20.5&50.0&59.1&32.8&57.7&55.8&3s&15.3\\
\rowcolor{green!40}\textbf{YOLOv7-521C}&\textbf{47.6}&\textbf{86.4}&\textbf{46.2}&\textbf{21.5}&44.9&\textbf{52.0}
&\textbf{21.2}&\textbf{51.5}&\textbf{59.2}&\textbf{36.1}&\textbf{58.1}&\textbf{62.5}&3s&15.3\\
YOLOv8\cite{yolov8}&61.2&92.8&67.7&18.3&59.5&59.9
&25.0&61.6&68.5&37.3&67.9&65.5&4s&20.5\\
\rowcolor{green!40}\textbf{YOLOv8-421C}&\textbf{61.6}&\textbf{93.2}&\textbf{68.9}&\textbf{22.6}&\textbf{61.1}&\textbf{60.9}
&24.9&61.7&\textbf{68.5}&\textbf{35.5}&\textbf{69.4}&\textbf{69.7}&4s&20.5\\
\rowcolor{green!40}\textbf{YOLOv8-521C}&61.1&92.9&68.5&14.7&60.5&60.1
&\textbf{25.1}&\textbf{62.0}&68.4&33.8&68.1&64.0&4s&20.5\\
\hline
\end{tabular}
\end{table*}

\subsubsection{Results on MS COCO TOD-80}
Table \ref{table_AP_KDN_tod} lists the performance of KDN (also known as 421C) on a single-branch detector compared to FPN-based multi-branch detectors and a NAS method on the MS COCO TOD-80 dataset. Table \ref{table_AP_coco_tiny} shows the AP and AR results of Cascade R-CNN-421C, YOLOv8-421C, and Swin Transformer-421C. From these results, we have the following observations.
\begin{itemize}
\item[i)]
In Table \ref{table_AP_KDN_tod}, we can observe that adding a Renormalized Connection to the single-branch detector Faster R-CNN (an implementation of KDN) greatly improves the performance of the 80-class small object detection task. KDN outperforms the listed multi-branch detectors. NASNet, on the other hand, was not able to converge in an acceptable time because even one step of training took a long time. This verifies that Renormalized Connection can cope with the multi-class scale-preferred problem.
\end{itemize}

\begin{figure}
\centering
\includegraphics[width=3.5in]{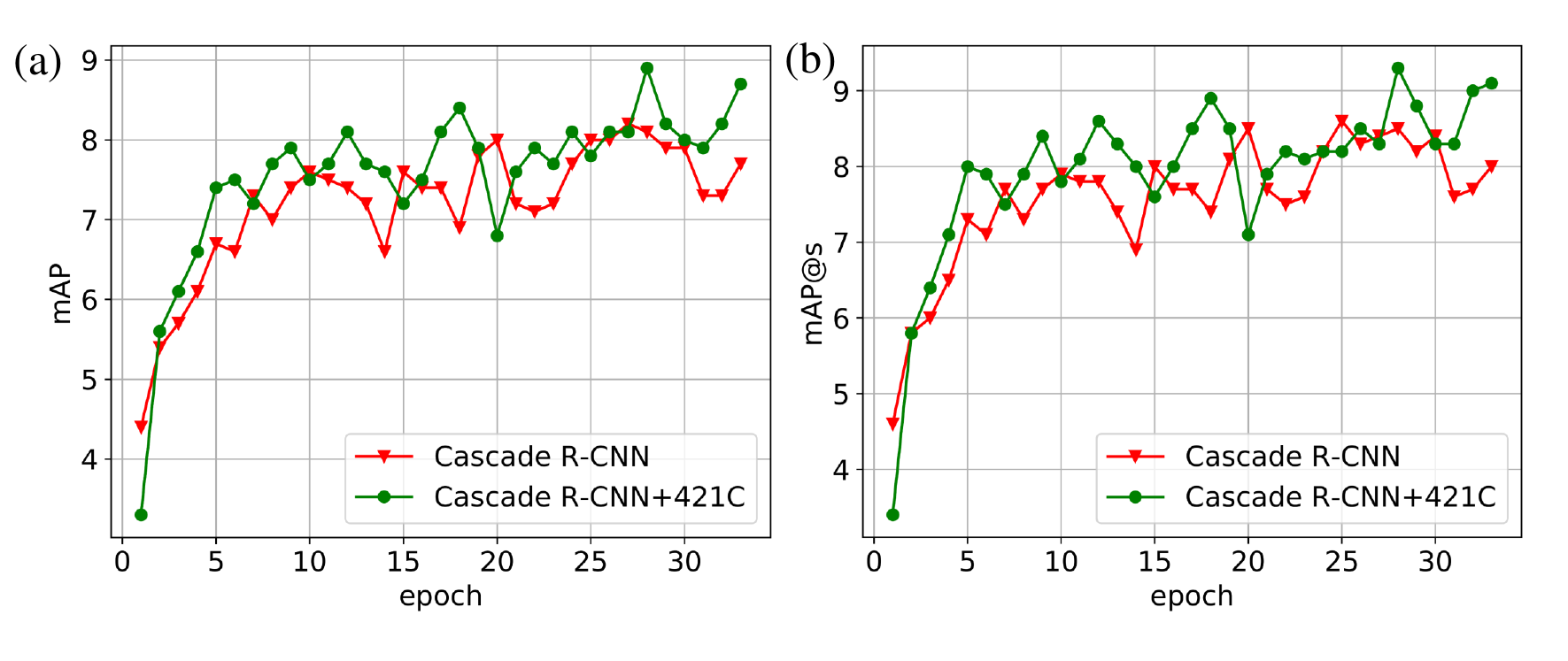}
\caption{MS COCO TOD-80 Val set inferential results of $AP^{0.5:0.95}$ (a) and $AP^{S}$ (b) plots of Cascade R-CNN-421C and baseline.} 
\label{coco_tiny_AP_plots}
\end{figure}

\begin{itemize}
\item[ii)]
As shown in Table \ref{table_AP_coco_tiny}, YOLOv8-421C achieves the highest AP of 11.4\% and $AP^{S}$ of 12.3\%. Cascade R-CNN-421C (8.9\%) beats the baseline (8.2\%) by 0.7\% at $AP^{0.5:0.95}$. It achieves $AP^{S}$ of 9.3\% with 0.9\% improvement compared to the baseline of 8.4\%, proving that 421C is effective for small object detection under complex scenes.
\end{itemize}

\begin{itemize}
\item[iii)]
Fig. \ref{coco_tiny_AP_plots} shows the average precision of Cascade R-CNN over the entire validation set (mainly considering $AP^{0.5:0.95}$ and $AP^{S}$) after each of the 33 epochs of training. The green line (421C) is almost always higher than the red line (baseline), suggesting that 421C brings a stable structured validity to the baseline. On multiclass complex natural scene scale-preferred datasets, 421C can effectively improve the prediction accuracy of small objects.
\end{itemize}

\subsubsection{Results on NWPU VHR-10}
The AP and AR comparisons of two typical $n21$C and baselines on NWPU VHR-10 are reported in Table \ref{table_APAR_nwpu}. The 421C further improves the performance of two real-time detectors. YOLOv7-421C outperforms YOLOv7 by 0.5\% in $AP^{0.5:0.95}$ and by 1.8\% in $AP^{S}$. YOLOv8-421C improves by 0.4\% in $AP^{0.5:0.95}$ and by 4.3\% in $AP^{S}$. During training, the highest $AP^{S}$ for YOLOv8 is 27\%, while YOLOv8-421C reaches 33.5\%. The 521C performs well on YOLOv7 but poorly on YOLOv8, suggesting that the doubled information obtained from the $P_3$ level in the scale-diversified task affects the performance of the anchor-based and anchor-free detectors differently. These results indicate that E421C performs well on the more complex multi-class arbitrary-sized satellite object detection dataset (the task with the most diverse scale distributions in this work).

\subsubsection{Ablation Study}
\begin{table*}
\renewcommand{\arraystretch}{1.5}
\setlength\tabcolsep{3pt}
\caption{Ablation study 1.1. AP and AR of YOLOv7-based Renormalized Connection with different connection strengths on IPIU test set.}
\label{table_APAR_ipiu_ablation}
\centering
\begin{tabular}{llllllllllll}
\hline
\rowcolor{gray!40}Method & $AP^{0.5:0.95}$ & $AP^{0.5}$ & $AP^{0.75}$ & $AP^{S}$ & $AR_{1}$ & $AR_{10}$ & $AR_{100}$ & $AR_{100}^{S}$ &T(s)${\downarrow}$ &T/f (ms) ${\downarrow}$\\
\hline
YOLOv7\cite{wang2022yolov7}&22.1&68.4&6.2&22.1&2.0&15.6&36.6&36.6&9s&16.88\\
\rowcolor{green!40}\textbf{YOLOv7-421C}&\textbf{28.7}&\textbf{78.5}&\textbf{12.0}&\textbf{28.7}&\textbf{2.5}&\textbf{19.3}&\textbf{41.6}&\textbf{41.6}&11s&20.63\\
\textbf{YOLOv7-521C}&27.7&77.0&11.0&27.7&2.3&18.6&41.6&41.6&9s&16.89\\
\textbf{YOLOv7-721C}&27.9&77.6&11.2&27.9&2.4&18.7&41.3&41.3&12s&22.51\\
\textbf{YOLOv7-821C}&26.1&74.8&9.7&26.1&2.3&18.0&40.0&40.0&11s&20.63\\
\textbf{YOLOv7-(3.1)21C}&22.7&68.2&7.1&22.7&2.1&16.2&36.9&36.9&12s&22.51\\
\textbf{YOLOv7-(3.5)21C}&28.2&77.9&11.6&28.2&2.5&19.0&41.4&41.4&12s&22.51\\
\hline
\end{tabular}
\end{table*}

\begin{table*}
\renewcommand{\arraystretch}{1.5}
\setlength\tabcolsep{3pt}
\caption{Ablation study 1.2. AP and AR of detectors with 421C, 521C, and baselines on RSOD Aircraft Val set. }
\label{table_AP_rsod_aircraft_ablation}
\centering
\begin{tabular}{lllllllllllllll}
\hline
\rowcolor{gray!40}Method & $AP$ & $AP^{0.5}$ & $AP^{0.75}$ & $AP^{S}$ & $AP^{M}$ & $AP^{L}$
& $AR_{1}$ & $AR_{10}$ & $AR_{100}$ & $AR_{100}^{S}$ & $AR_{100}^{M}$&$AR_{100}^{L}$
&T (s) ${\downarrow}$&T/f (ms) ${\downarrow}$\\
\hline
YOLOv7\cite{wang2022yolov7}&66.4&97.1&81.2&44.5&\textbf{68.4}&79.0
&6.6&\textbf{48.6}&\textbf{72.3}&\textbf{58.4}&73.8&\textbf{83.8}&1s&13.15\\
\rowcolor{green!40}\textbf{YOLOv7-421C}&\textbf{66.5}&\textbf{97.2}&\textbf{82.0}&\textbf{45.4}&68.2&\textbf{79.0}
&\textbf{6.6}&48.4&72.2&57.6&\textbf{73.9}&82.3&\textbf{1s}&13.15\\
\rowcolor{green!40}\textbf{YOLOv7-521C}&65.1&96.9&78.6&41.6&67.5&77.0
&6.1&47.7&71.2&55.4&73.2&81.1&\textbf{1s}&13.15\\
YOLOv8\cite{yolov8}&71.0&\textbf{97.1}&\textbf{88.1}&\textbf{52.3}&72.2&82.9
&\textbf{6.8}&50.8&\textbf{76.0}&\textbf{62.1}&\textbf{77.6}&86.7
&1s&13.15\\
\rowcolor{green!40}\textbf{YOLOv8-421C(A)}&\textbf{71.5}&\textbf{97.1}&87.4&51.7&72.6&\textbf{84.6}
&\textbf{6.8}&\textbf{51.1}&75.8&61.1&77.4&\textbf{87.3}
&\textbf{1s}&13.15\\
\rowcolor{green!40}\textbf{YOLOv8-421C(B)}&\textbf{71.5}&\textbf{97.1}&86.3&52.1&\textbf{73.1}&82.6
&6.7&51.0&75.9&61.6&77.5&86.3
&\textbf{1s}&13.15\\
\rowcolor{green!40}\textbf{YOLOv8-521C}&71.3&\textbf{97.1}&87.7&51.6&\textbf{73.1}&82.1
&\textbf{6.8}&51.0&75.8&61.2&\textbf{77.6}&86.3&\textbf{1s}&13.15\\
\hline
\end{tabular}
\end{table*}

\begin{table*}
\renewcommand{\arraystretch}{1.5}
\setlength\tabcolsep{3pt}
\caption{Ablation study 1.3. AP and AR of detectors with 421C, 521C, and baselines on RSOD Val set.}
\label{table_AP_rsod_ablation}
\centering
\begin{tabular}{lllllllllllllll}
\hline
\rowcolor{gray!40}Method & $AP$ & $AP^{0.5}$ & $AP^{0.75}$ & $AP^{S}$ & $AP^{M}$ & $AP^{L}$& $AR_{1}$ & $AR_{10}$ & $AR_{100}$ & $AR_{100}^{S}$ & $AR_{100}^{M}$&$AR_{100}^{L}$ &T(s) ${\downarrow}$&T/f (ms) ${\downarrow}$\\
\hline
YOLOv7\cite{wang2022yolov7}&62.7&90.4&70.9&50.7&71.8&66.3
&31.2&60.5&70.8&61.1&79.0&73.7&4s&25.31\\
\rowcolor{green!40}\textbf{YOLOv7-421C}&62.9&91.4&72.1&48.6&71.8&66.4
&32.3&60.7&71.4&59.5&78.8&74.0&3s&18.98\\
\rowcolor{green!40}\textbf{YOLOv7-521C}
&63.0&90.7&71.3&49.6&72.6&66.5
&32.7&60.7&71.4&59.9&78.2&74.3&3s&18.98\\
YOLOv8\cite{yolov8}&\emph{68.4}&93.1&77.9&50.1&72.5&71.7
&36.0&65.1&75.7&60.1&78.5&78.6&4s&25.31\\
\rowcolor{green!40}\textbf{YOLOv8-421C}&\textbf{68.7}&93.5&79.6&49.6&74.3&71.8&36.2&64.2&74.4&60.3&79.7&77.2&4s&25.31\\
\rowcolor{green!40}\textbf{YOLOv8-521C}&68.7&93.6&76.5&50.4&73.9&72.4
&35.6&64.8&75.2&60.3&80.0&78.4&4s&25.31\\
YOLOv8 Top$AP^{S}$&67.0&92.1&76.7&\emph{51.9}&73.8&70.0&34.6&64.4&75.4&\emph{61.7}&79.4&78.0&4s&25.31\\
\rowcolor{green!40}\textbf{YOLOv8-421C} Top$AP^{S}$&67.3&91.9&75.0&\textbf{52.6}&74.9&70.7&34.6&64.3&75.4&\textbf{62.4}&79.7&78.3&4s&25.31\\
\rowcolor{green!40}\textbf{YOLOv8-521C} Top$AP^{S}$&
68.7&92.5&77.4&51.8&74.9&72.3
&35.6&65.6&76.4&61.0&79.8&79.7&4s&25.31\\
\hline
\end{tabular}
\end{table*}

To gain a deeper insight into Renormalized Connections (RCs), we continue to perform ablation experiments on satellite image datasets. Firstly, we conduct a series of RC experiments on YOLOv7 with different connection strengths to investigate the renormalization effect of the uniform form 421C and other connections, as shown in Table \ref{table_APAR_ipiu_ablation} -- Table \ref{table_AP_rsod_ablation} and Table \ref{table_APAR_nwpu}. Secondly, we add three representative modules to the linear $n21$C and show their performance separately. Thirdly, we compare two variants of E421C that take as input some of the features in the feature cascade, viz: 1) $\text{\textit{421C}}^{s\&m}$ and 2) $\text{\textit{421C}}^{s\&l}$. 

\noindent\textbf{Ablation study 1}
In KDN, we use the amplification factor $\lambda$ to describe the renormalization rate of all features in the feature cascade. Similarly, in the generalized Renormalized Connection $n21$C used in multi-branch detector, we use connection strength to represent the same description of the amplification factor. In the scale-preferred task, we only consider adjusting the connection strength at the $P_3$ level.
\begin{itemize}
\item[1.1)]
As shown in Table \ref{table_APAR_ipiu_ablation}, we examined six amplified or deamplified factors compared to the 421 factors in the difficult scale-preferred tiny object detection task. The uniform connection strength of RC (421C) is superior to that of the other types of RCs. This study demonstrates that the simplest E421C that satisfies the scaling property of renormalization group theory is suitable for solving hard scale-preferred problems as described in Section \ref{renorm} and \ref{insights}.
\end{itemize}

\begin{itemize}
\item[1.2)]
In the RSOD Aircraft experiments as in Table \ref{table_AP_rsod_aircraft_ablation}, we compare the performance of two typical $n21$Cs, the 421C, and the 521C, on a small object detection task (with large scale range compared to tiny objects). The YOLOv7-521C outperforms the 421C on mAP but performs poorly on $AP^{S}$ and most ARs. However, the 521C performed worse than 421C on YOLOv8. YOLOv7's results on detecting small/medium-sized objects differ from the results on the difficult scale-preferred IPIU task.
\end{itemize}

\begin{itemize}
\item[1.3)]
As shown in Table \ref{table_AP_rsod_ablation}, the performance of YOLOv8-521C is comparable to that of 421C, but with a higher AP for small objects. For YOLOv7, the performance of 521C is slightly better than that of 421C. This diverse scale-distributed dataset has similar performance for different types of Renormalized Connections. 
\end{itemize}

\begin{itemize}
\item[1.4)]
On the most diverse scale distributed NWPU dataset, 421C, and 521C perform better on YOLOv8 and YOLOv7, respectively, as shown in Table \ref{table_APAR_nwpu}. When performing YOLOv7-RC on scale-diversified datasets, the larger connection strength of $P_3$ can promote the $AP^S$, which is consistent with the general intuition of weighted feature fusion. This further confirms that RC can bias the detector towards the objective of its embedded branch.
\end{itemize}

\begin{table*}
\renewcommand{\arraystretch}{1.5}
\setlength\tabcolsep{3pt}
\caption{Ablation study 2. AP and AR of YOLOv7 based n21C added with extra representative modules on IPIU test set. $1\times1$ convolution layer, norm or bn, activation: ReLu. The best AP, AR, and T are in bold.}
\label{table_APAR_ipiu_n21c}
\centering
\begin{tabular}{lllllllllllllll}
\hline
\rowcolor{gray!40}Method &$1\times1$ conv&norm&act& $AP^{0.5:0.95}$ & $AP^{0.5}$ & $AP^{0.75}$ & $AP^{S}$ & $AR_{1}$ & $AR_{10}$ & $AR_{100}$ & $AR_{100}^{S}$ &T(s)${\downarrow}$ &T/f (ms) ${\downarrow}$\\
\hline
YOLOv7\cite{wang2022yolov7}&&&&22.1&68.4&6.2&22.1&2.0&15.6&36.6&36.6&9s&16.88\\
\rowcolor{green!40}\textbf{YOLOv7-421C}&&&&\textbf{28.7}&\textbf{78.5}&12.0&\textbf{28.7}&\textbf{2.5}&\textbf{19.3}&41.6&41.6&11s&20.63\\
\textbf{YOLOv7-521C}&&&&27.7&77.0&11.0&27.7&2.3&18.6&41.6&41.6&9s&16.89\\
\textbf{YOLOv7-521C}&\checkmark&&&28.2&77.6&11.8&28.2&2.4&18.8&41.8&41.8&11s&20.63\\
\textbf{YOLOv7-421C}&&\checkmark&&27.7&77.4&10.7&27.7&2.4&18.8&41.0&41.0&12s&22.51\\
\textbf{YOLOv7-421C}&&&\checkmark&\textbf{28.7}&\textbf{78.5}&\textbf{12.2}&\textbf{28.7}&2.4&19.1&\textbf{42.0}&\textbf{42.0}&11s&20.63\\
\textbf{YOLOv7-421C}&\checkmark&\checkmark&\checkmark
&23.6&71.6&6.5&23.6&2.0&16.6&38.1&38.1&12s&22.51\\
\hline
\end{tabular}
\end{table*}

\noindent\textbf{Ablation study 2}
We conducted a series of experiments on the IPIU test set by adding common modules used to enhance the representation capability on the linear $n21$C to investigate further potential factors affecting the performance gain. As shown in Table \ref{table_APAR_ipiu_n21c}, the performance of YOLOv7-421C with the added ReLU activation function is comparable to that of the initial linear 421C. The other modules do not improve the performance over the linear $n21$C but make it worse. This verifies that the linear Renormalized Connections can handle the scale-preferred problem well, even under extreme conditions such as purely tiny object detection in satellite images.

\noindent\textbf{Ablation study 3}
The $\text{\textit{421C}}^{s\&m}$ is a partial branch connected network where $F^s$ and $F^m$ pass through the 421 Connection to enhance the output of $\text{\textit{421C}}^s$. Compared to 421C, the large object branch $F^l$ does not participate in the detection of small objects in $\text{\textit{421C}}^{s\&m}$. Similarly, the medium object branch $F^m$ is absent from the connection operation of $\text{\textit{421C}}^{s\&l}$. The mAP and AR results of 421C and its variants are given in Table \ref{table_ablation_ipiu}. From this table, we can see the following.
\begin{itemize}
\item[a)]
$\text{\textit{421C}}^{s\&m}$ and $\text{\textit{421C}}^{s\&l}$ have some performance improvements over the baseline, indicating that the 421 Connection is useful for small object detection.
\end{itemize}
\begin{itemize}
\item[b)]
421C outperforms $\text{\textit{421C}}^{s\&m}$ and $\text{\textit{421C}}^{s\&l}$. These results demonstrate that both medium and large object branches' features are crucial for a complete description of the feature in the small object branch.
\end{itemize}
\begin{itemize}
\item[c)]
$\text{\textit{421C}}^{s\&m}$ and $\text{\textit{421C}}^{s\&l}$ achieve similar performance regardless of the choice of detection branch, suggesting that 421 Connection is suitable for effective feature enhancement with fewer branches.
\end{itemize}

\begin{table}
\renewcommand{\arraystretch}{1.5}
\setlength\tabcolsep{1pt}
\caption{Ablation study 3. YOLOv8-421C and its variants on the IPIU test.}
\label{table_ablation_ipiu}
\centering
\begin{tabular}{lllllllllll}
\hline
\rowcolor{gray!40}Method & $AP$ & $AP^{0.5}$ & $AP^{0.75}$ & $AP^{S}$ & $AR_{1}$ & $AR_{10}$ & $AR_{100}$ & $AR_{100}^{S}$ &T(s)${\downarrow}$ \\
\hline
$\text{\textit{421C}}^{s\&m}$ &55.4&94.2&57.0&55.4&3.6&30.7&64.2&64.2&8s \\
$\text{\textit{421C}}^{s\&l}$
&55.4&94.1&57.2&55.4&3.6&30.6&64.3&64.3&9s\\
\rowcolor{green!40}421C&\textbf{55.5}&\textbf{94.2}&\textbf{57.6}&\textbf{55.5}&\textbf{3.6}&\textbf{30.9}&\textbf{64.4}&\textbf{64.4}&9s\\
\hline
\end{tabular}
\end{table}

\subsubsection{Analysis on Different Sizes}
We will discuss how the proposed method affects the AP and AR at different object sizes. For full-size detection benchmarks, as shown in Fig. \ref{rsod_AP_plots}, 421C not only achieves higher mAP but also improves the AP for small objects. In the last two rows of Table \ref{table_AP_rsod}, YOLOv8-421C increases the $AP^{0.5:0.95}$, $AP^{S}$, $AP^{M}$ and $AP^{L}$ by 0.3\%, 0.7\%, 1.1\%, and 0.7\%, respectively; and increases the $AR_{100}^{S}$, $AR_{100}^{M}$ and $AR_{100}^{L}$ by 0.7\%, 0.3\%, 0.3\% respectively. In summary, 421C improves the detection performance of small, medium, and large objects.

\begin{figure*}
\centering
\includegraphics[width=7.16in]{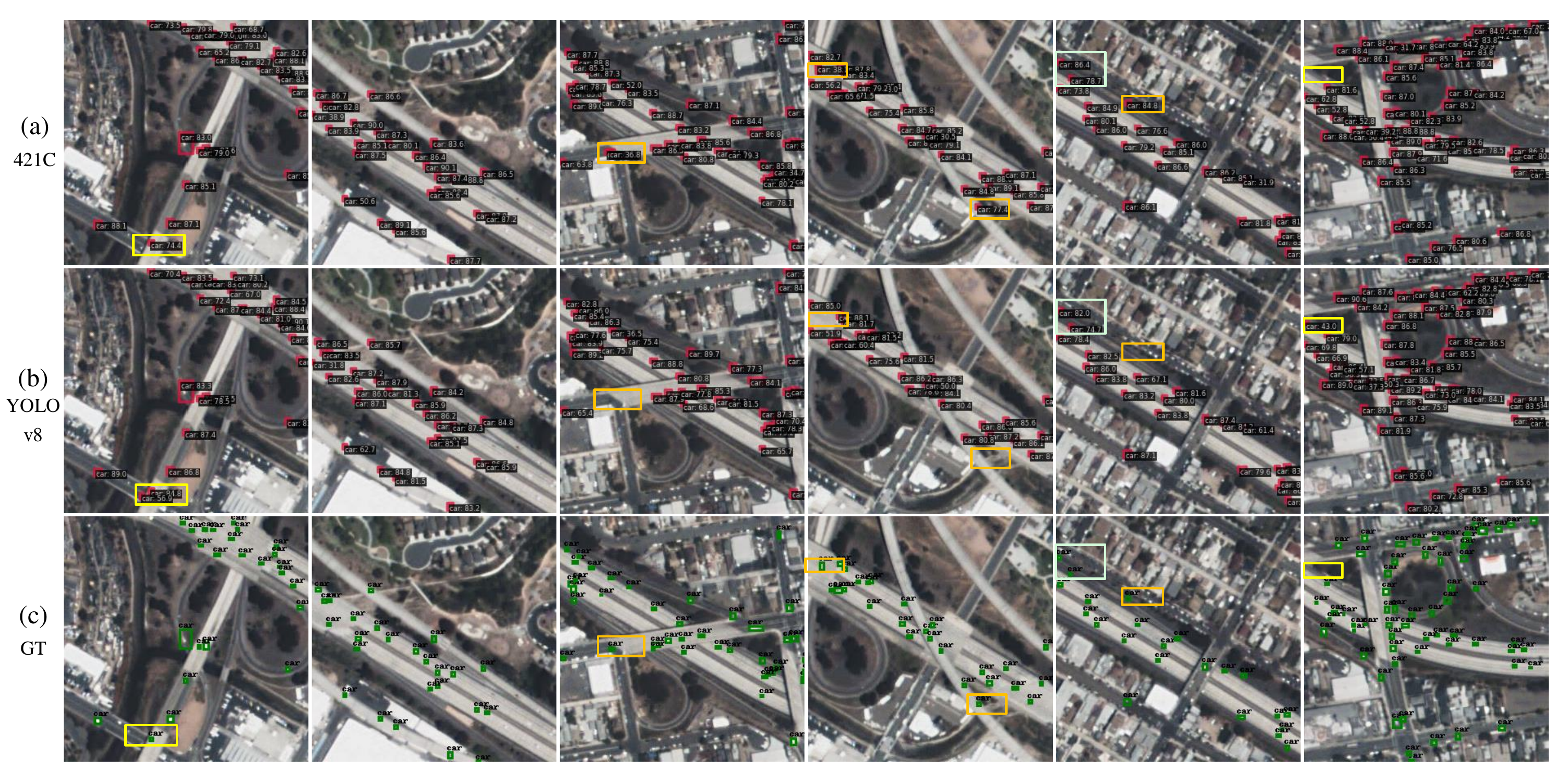}
\caption{Visualized results of (a) YOLOv8-421C, (b) YOLOv8, and (c) GT on IPIU test set. Yellow boxes indicate false positives for YOLOv8. Orange boxes indicate missing detections for YOLOv8. The Wathet box indicates dark backgrounds but accurately detects results.}
\label{ipiu_visualized_results}
\end{figure*}
\begin{figure*}
\centering
\includegraphics[width=7.16in]{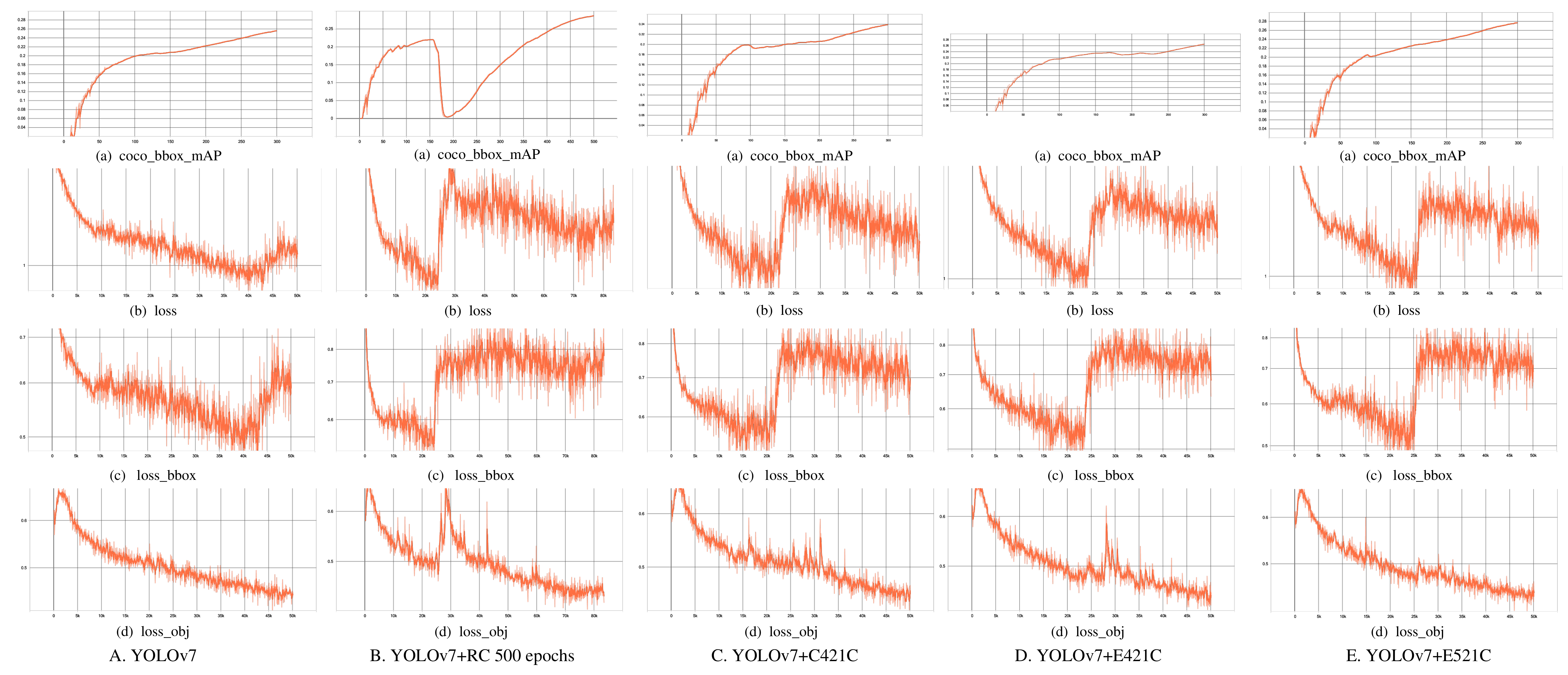}
\caption{AP and loss curves of YOLOv7 (baseline), Renormalized Connection, Complete 421C, Economical 421C, and Economical 521C on IPIU Val set.}
\label{yolov7_curves}
\end{figure*}
\subsubsection{Visualization of Predicted Results}
To have some visualized observations, we further show the visualized results of YOLOv8-421C on the IPIU test set in Fig. \ref{ipiu_visualized_results}. In Fig. \ref{ipiu_visualized_results}, the three images in each column represent the results of YOLOv8-421C, YOLOv8, and GT on one image. We manually encircled the false detections, missing detections, and dark background detection results with yellow, orange, and wathet boxes, respectively. Compared to YOLOv8, YOLOv8-421C predicts moving vehicles in complex scenes more accurately in the following aspects. 1) YOLOv8-421C has higher mAP and single instance AP than YOLOv8. 2) Vehicles can be accurately detected even in dark backgrounds. 3) The 421C mitigates the problems of false detection and missing detection. 4) It correctly identifies the desired objects and ignores other similar objects such as vehicles in the parking lots. From these visualization results, we can see that YOLOv8-421C can cope with the hard scale-preferred tasks in scenes that are blurred, low contrast, and full of interfering objects.

\subsubsection{Comparison of AP and Loss Curves}
In Fig. \ref{yolov7_curves}, we show the mAP and loss curves of the five YOLOv7-based detectors on the IPIU dataset. The left column shows the YOLOv7 baseline using PAFPN as the feature extractor. In the second column, we can see that the accuracy of PAFPN with RC decreases rapidly during training, but gains greater acceleration after the mAP value drops to 0 and reaches a higher value at the end of training. The loss curves of different types of $n21$C have a similar trend, i.e., the total loss continues to decrease after a step change, while the baseline saturates at last. RC has the property of bottoming out quickly and recovering faster, unlike conventional deep network training. The phenomenon of RC (the 2nd column) resembles the ``hit bottom and rebound''. It is able to rebound from a brief crisis. 

The AP curves further demonstrate that the Renormalized Connections do rearrange the multi-level features of the feature extractor during the training process and learn in a more correct direction, ultimately achieving better experimental results.

\section{Conclusion}
\label{conclusion}
In this article, we studied the challenging problem of scale-preferred detection in satellite imagery, especially for the difficult tiny object detection. Inspired by the renormalization group theory, we design a Knowledge Discovery Network to implement the theory on efficient feature extraction. According to the observations of KDN experiments, we abstract a class of renormalized connections with different connection strengths called $n21$C. 

After conducting experiments on scale-preferred tasks of the original YOLOv8-PAFPN, we observed that the ``divide-and-conquer'' design ideas of the FPN severely hampers detector learning due to the presence of a large number of negative samples at other scales as well as interfering activations from background noise. These actively generated negative samples cannot be eliminated by the focal loss function. Therefore, we need to build a bridge between the multi-level extractor and the multi-branch head network to renormalize the information flow and achieve the reorientation of the learning process. Naturally, we can implement this bridge with the help of Renormalized Connections abstracted from KDN. 

We then generalize the RCs to FPN-based multi-branch detectors and verify that this approach solves the above problem well. The $n21$C not only extends the multi-level and ``divide-and-conquer'' mechanism of the multi-level features of the FPN-based detectors to a wide range of scale-preferred tasks but also enables synergistic effects of multi-level features on the specific learning objectives. The $n21$Cs assign appropriate subtasks to multiple detection branches for scale-preferred tasks. In addition, interfering activations, including those from the background and interfering negative samples, are greatly reduced and the detector redirects the learning direction to the correct one. 

The Renormalized Connection renormalizes the information flow in all phases of network learning, rearranging the feature flow in the forward propagation phase and reorienting the gradient flow in the backpropagation phase. The information flow in both phases together determines the learning direction of the network. The design mechanism of Renormalized Connection lies in its ``synergistic effect'' on multi-scale features and its ``focus'' on the dominant objective.

To meet different task requirements, the Renormalized Connections can be seen as a customized plug-in module to renormalize the information flow of the network and redirect the learning process. The customized plug-in property means that there is flexibility in designing different connection strengths, embedding positions (between the feature extractor and the head in which detection branch or branches), and the number of Renormalized Connections so that the ``synergistic focusing'' mechanism can be integrated into the detector and adapted to different task requirements. 

Extensive experiments of 17 well-designed detection architectures embedded with $n21$Cs on five different levels of scale-preferred tasks (including a newly released high-difficulty scale-preferred dataset and scale-diversified tasks) validate the renormalization effect of the Renormalized Connection approach $n21$Cs, especially the simplest linear form E421C. And it achieves real-time detection without additional parameters, but with higher accuracy. We hope that this lightweight and generalizable Renormalized Connection will bring some improvements to scale-preferred object detection research and applications. 

The learning process of Renormalized Connection differs from the loss reduction and optimization process of existing deep learning networks. In future work, we will further deepen the training process of $n21$Cs on small object detection tasks. Besides, we can continue to improve the representational capacity of the detector to further enhance its performance.

\section*{Acknowledgments}

The authors thank Zhaoliang Pi for his help in annotating the IPIU source dataset. This work pays homage to the excellent ResNet, FPN, PAFPN, HTC, YOLOv7, YOLOv8, and so on. The authors would like to thank the Editor, the Associate Editor, and the anonymous reviewers for their constructive comments and suggestions.


%

\bibliographystyle{IEEEtran}
\bibliography{Bibliography_file.bib}


\vfill

\end{document}